\documentclass{article} 
\usepackage{iclr2027_conference,times}

\usepackage{amsmath,amsfonts,bm}

\def\eqref#1{equation~\ref{#1}}

\def\1{\bm{1}}

\DeclareMathAlphabet{\mathsfit}{\encodingdefault}{\sfdefault}{m}{sl}
\SetMathAlphabet{\mathsfit}{bold}{\encodingdefault}{\sfdefault}{bx}{n}

\usepackage{hyperref}
\usepackage{url}
\usepackage{booktabs}
\usepackage{graphicx}
\usepackage{comment}
\usepackage{color}
\usepackage{colortbl}
\usepackage{xcolor}
\usepackage{xspace}
\usepackage{enumitem}
\usepackage{algorithm,algpseudocode}

\usepackage{makecell}
\usepackage{threeparttable}
\usepackage{amssymb}

\title{MIRAGE: Metric Inference and Reliability-Aware Gaussian Estimation from Feed-Forward Vision}

\author{
Xinzhe Wang$^{1,2}$\footnotemark[1] \quad
Changjian Jiang$^{3,2}$\footnotemark[1] \quad
Kaiwen Song$^{4,2}$ \quad
Xudong Li$^{5,2}$ \quad
Kerui Ren$^{1,2}$ \quad \\
\textbf{Ran Yi}$^{1}$\footnotemark[2] \quad
\textbf{Lizhuang Ma}$^{1}$\footnotemark[2] \quad
\textbf{Chunhua Shen}$^{6,2}$ \quad
\textbf{Linning Xu}$^{7,2}$ \quad
\textbf{Tao Lu}$^{2}$ \quad
\textbf{Mulin Yu}$^{2}$\footnotemark[3] \quad \\[0.5em]
{\small $^1$Shanghai Jiao Tong University,
$^2$Shanghai Artificial Intelligence Laboratory,} \\
{\small $^3$The University of Hong Kong,
$^4$University of Science and Technology of China,} \\
{\small $^5$Fudan University,
$^6$Zhejiang University,
$^7$The Chinese University of Hong Kong}
}

\iclrfinalcopy 
\begin{document}

\begingroup
\renewcommand{\thefootnote}{\fnsymbol{footnote}}
\maketitle
\let\thefootnote\relax
\footnotetext{%
  $^{*}$Denotes joint contribution.\quad
  $^{\dagger}$Corresponding authors.\quad
  $^{\ddagger}$Project leader.%
}
\endgroup

\begin{figure}[!ht]
    \centering
    \includegraphics[width=\linewidth]{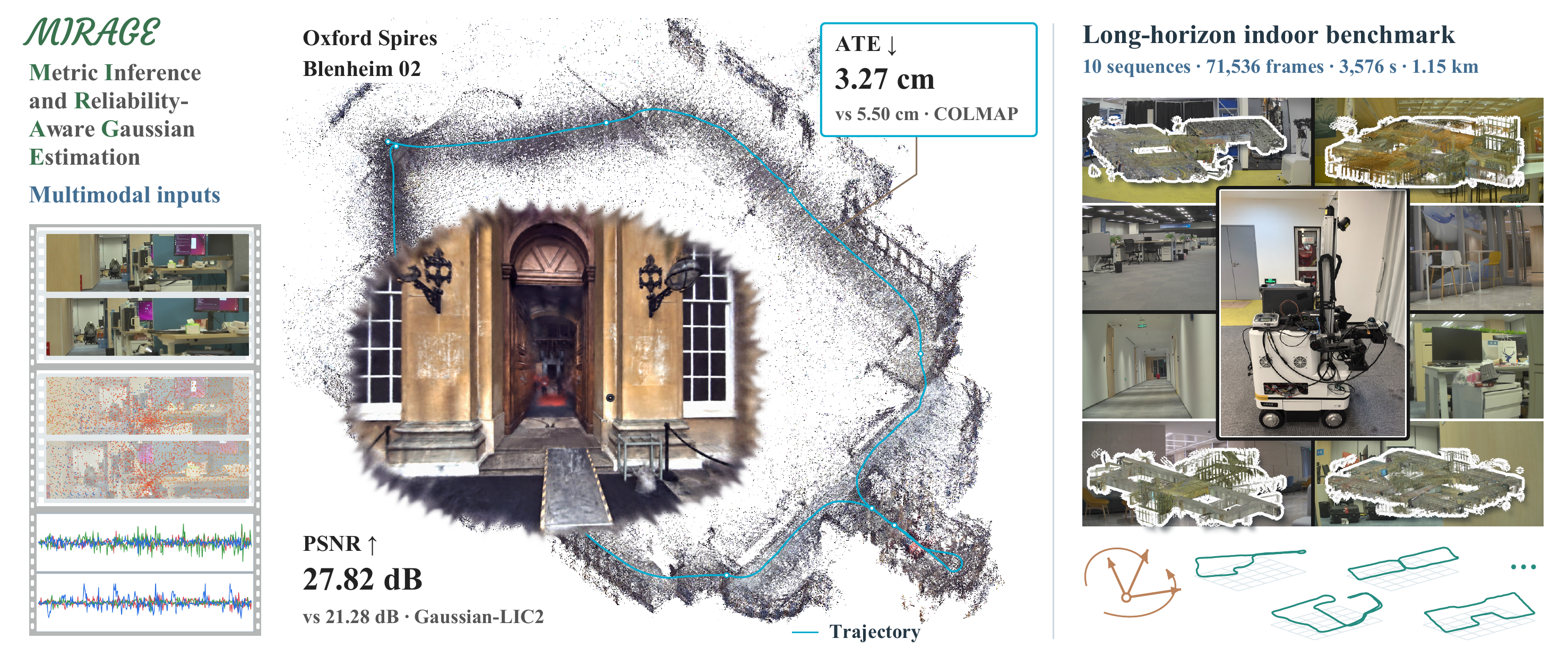}
    \caption{We propose \textbf{MIRAGE}, which anchors feed-forward visual geometry to LiDAR--IMU motion through reliability-aware pose refinement. \textbf{Left:} on Oxford Spires Blenheim~02, MIRAGE achieves lower ATE than COLMAP (3.27 vs.\ 5.50~cm) and higher rendering PSNR than Gaussian-LIC2 (27.82 vs.\ 21.28~dB). \textbf{Right:} our synchronized multimodal platform supports a long-horizon indoor benchmark with 10 sequences and MIRAGE-established 6-DoF accurate reference trajectories for evaluating vision-only reconstruction. 
    Project Page: \color{magenta}\textbf{\href{https://hiroxzwang.github.io/publications/mirage/}{MIRAGE-URL}}.}
    \label{fig:mirage_teaser}
\end{figure}

\begin{abstract}
{Feed-forward 3D priors enable high-quality reconstruction from image streams but often lack reliable metric scale and suffer from long-term trajectory drift. In contrast, LiDAR--inertial estimation provides metrically accurate motion but lacks the image-space consistency needed for high-fidelity reconstruction. We propose \textbf{MIRAGE}, which anchors feed-forward visual geometry to LiDAR--IMU metric motion and regulates online pose corrections according to correspondence support and cross-modal agreement, jointly enabling accurate metric pose estimation and high-fidelity Gaussian reconstruction. Experiments on  Oxford Spires, M2DGR and INS show that MIRAGE achieves accurate metric trajectory estimation, outperforming COLMAP, a widely used but computationally expensive reference for vision-only benchmarks. Motivated by the scarcity of long-horizon monocular benchmarks for large-scale indoor reconstruction, we further build a multimodal data-collection platform and use MIRAGE to establish accurate 6-DoF reference trajectories for a new vision-only benchmark. We benchmark recent monocular large-scale reconstruction methods on this dataset, providing a systematic evaluation of their performance in long-horizon indoor scenes.}
\end{abstract}

\section{Introduction}
{Recent advances in 3D scene representation and camera pose estimation have substantially improved the quality and practicality of image-based reconstruction. 3D Gaussian Splatting (3DGS) \citep{kerbl2023gaussian} represents scenes with optimizable anisotropic Gaussian primitives and an efficient differentiable rasterizer, enabling high-fidelity real-time novel-view synthesis. In parallel, feed-forward geometric models~\citep{leroy2024grounding,wang2025vggt,wang2025pi} have made substantial progress in jointly inferring camera poses and scene geometry from unposed images. These advances have further driven the development of increasingly capable online tracking, mapping, and on-the-fly reconstruction systems~\citep{matsuki2024gaussianslam,huang2024photoslam,keetha2024splatam,meuleman2025onthefly,wu2025monocular}. More recent methods combine SLAM frameworks with data-driven feed-forward priors to further improve pose estimation and reconstruction quality in streaming scenarios~\citep{li2026artdeco,ren2026m}. However, purely visual reconstruction remains subject to global similarity ambiguity, while accumulated errors in learned geometry and pose estimation can lead to scale and motion drift over long sequences. These limitations highlight the need for an explicit metric anchor to achieve both high-fidelity reconstruction and a physically grounded trajectory.
}

{LiDAR--IMU sensing provides the metric scale and motion constraints missing from purely visual reconstruction. However, simply introducing a metric anchor does not resolve how feed-forward visual geometry should influence online pose refinement. Visual priors can improve local alignment and reconstruction quality, yet unreliable corrections may compromise metric consistency, while overly suppressing them can forfeit useful image-space refinement. Since the reliability of visual corrections varies with correspondence support and cross-modal agreement, a fixed fusion strategy is insufficient. The key challenge is therefore to selectively exploit feed-forward visual geometry while preserving a metrically consistent trajectory.
}

{We propose \textbf{MIRAGE}\footnote{The name reflects the gap between visual plausibility and metric consistency.}, an on-the-fly reconstruction framework that combines feed-forward visual geometry with LiDAR--IMU sensing to jointly achieve metrically accurate pose estimation and high-fidelity Gaussian reconstruction. MIRAGE uses LiDAR--IMU motion as a metric anchor and selectively incorporates visual pose corrections according to correspondence support and cross-modal agreement, retaining useful visual refinement without compromising metric consistency. Given the stabilized metric trajectory, MIRAGE further leverages image residuals and LiDAR geometry for structure-aware Gaussian refinement, improving reconstruction quality while keeping the camera poses fixed. Experiments on Oxford Spires~\citep{tao2026oxfordspires}, M2DGR~\citep{yin2022m2dgr} and INS~\citep{deng2025mneslam} demonstrate that MIRAGE achieves accurate and metrically consistent trajectory estimation while maintaining high-fidelity, photorealistic Gaussian reconstruction. Notably, MIRAGE achieves higher trajectory accuracy than COLMAP, despite COLMAP being a computationally expensive method widely used to establish reference trajectories for vision-only benchmarks. This demonstrates MIRAGE's potential to provide reliable reference trajectories for benchmarking vision-only reconstruction methods.}


{Long-horizon benchmarks for monocular reconstruction in large-scale indoor environments remain limited. To fill this gap, we develop a remotely operated capture platform with hardware-synchronized cameras, LiDAR, and IMU, and collect ten long-horizon sequences across diverse indoor scenes, as shown in Figure~\ref{fig:mirage_teaser}. MIRAGE is then used to establish accurate 6-DoF reference trajectories for these sequences, forming a vision-only benchmark for large-scale indoor reconstruction. We further evaluate recent long-horizon monocular reconstruction methods on this benchmark.}

Our contributions are threefold:
\begin{itemize}[leftmargin=1.2em]
    \item 
    {
    \textbf{Reliability-gated metric trajectory inference.} We anchor feed-forward visual geometry to LiDAR--IMU metric motion and treat visual pose updates as correction proposals, selectively committing them according to correspondence support and cross-modal consistency to improve image-space alignment without compromising metric trajectory accuracy.
    }
    \item \textbf{Pose-fixed structure-aware Gaussian refinement.} 
    {
    We decouple scene refinement from trajectory estimation by fixing the metric camera poses and using image residuals together with LiDAR depth evidence to diagnose and repair oversized, misplaced, and missing primitives, improving geometric detail and appearance without perturbing the recovered trajectory.
    }
    \item {
    \textbf{A long-horizon benchmark dataset for real-world indoor reconstruction.} We introduce a large-scale indoor benchmark comprising 10 synchronized RGB--LiDAR--IMU sequences, with 71,536 RGB frames spanning 59.6 minutes and 1.15 km across laboratories, dining areas, offices, and long corridors. The benchmark provides full 6-DoF reference trajectories and supports joint evaluation of long-horizon metric localization and high-fidelity 3D reconstruction.
    }
\end{itemize}

\section{Related Work}
\label{sec:related_work}

\subsection{Visual Pose Estimation and Streaming Gaussian Reconstruction}
\label{sec:related_work_visual}

{Learned geometry models~\citep{wang2024dust3r,leroy2024grounding,wang2025vggt,lin2025depth} provide dense priors for pose estimation and reconstruction. Recent methods incorporate these priors into visual tracking and global optimization~\citep{murai2025mast3rslam,maggio2026vggt,maggio2026vggt2}, exploit geometric context for reconstruction~\citep{chen2026lingbotmap,xie2026scal3r}, or combine feed-forward predictions with global structure-from-motion optimization~\citep{deng2026glob3r,pan2026gluemap}.}

{Gaussian SLAM~\citep{matsuki2024gaussianslam,huang2024photoslam,keetha2024splatam} extends 3DGS~\citep{kerbl2023gaussian} to incremental tracking and mapping. Other approaches predict Gaussians from image streams~\citep{chen2024pref3r,li2025streamgs}, optimize scenes incrementally~\citep{meuleman2025onthefly,meuleman2026immediate3dgs}, or combine learned priors with online pose and map optimization~\citep{lin2025longsplat,li2026artdeco,jiang2026planing,ren2026m}. Monocular reconstruction nevertheless remains scale-ambiguous, and learned priors do not guarantee metric accuracy or drift-free trajectories over long traversals.}

\subsection{LiDAR--Inertial Estimation and Multimodal Gaussian Mapping}
\label{sec:related_work_multimodal}

{LiDAR--inertial estimators~\citep{shan2020liosam,xu2022fastlio2,liu2026voxelslam} recover metric motion from range and inertial constraints, while visual--LiDAR systems~\citep{shan2021lvisam,lin2022r3live,zheng2022fast,zheng2025fastlivo2,zhao2025resilient} incorporate image observations. Multimodal Gaussian mapping~\citep{hong2024livgaussmap,lang2025gaussianlic,xie2025gslivm,jiang2024li,deng2025vpgs,cui2024letsgo} uses metric poses and LiDAR geometry for initialization and photometric refinement. Recent methods further incorporate Gaussian-map observations into localization~\citep{hong2025gslivo,lang2026gaussianlic2} or feed-forward priors into estimation and mapping~\citep{wang2026lidarvggt,li2026rmgsslam}. MIRAGE regulates pose corrections using visual and geometric evidence to support both metric accuracy and rendering quality.}



\subsection{Benchmarks for Multimodal Localization and Reconstruction}
\label{sec:related_work_benchmarks}

{Visual benchmarks~\citep{sturm2012tumrgbd,dai2017scannet,knapitsch2017tanks} support tracking or reconstruction evaluation without mobile 3D LiDAR streams. ScanNet++~\citep{yeshwanth2023scannetpp} combines high-resolution RGB, surveyed laser geometry, and mobile RGB-D/IMU data. Mobile datasets vary in sensing and supervision: MulRan~\citep{kim2020mulran} lacks camera images, Hilti-Oxford~\citep{zhang2023hiltioxford} uses grayscale cameras, and FAST-LIVO2's self-collected sequences~\citep{zheng2025fastlivo2} lack pose ground truth. RGB--LiDAR--IMU datasets~\citep{tao2026oxfordspires,yin2022m2dgr,jiao2022fusionportable,wei2025fusionportablev2,zhao2024subtmrs} cover diverse environments and platforms, while recent multimodal benchmarks~\citep{deng2025mneslam,lang2026gaussianlic2} provide reference geometry and poses for localization and rendering evaluation. Nevertheless, synchronized RGB--LiDAR--IMU benchmarks with full 6-DoF reference trajectories over long, object-rich indoor traversals remain scarce. Our extended captures in laboratories, offices, and dining spaces address this gap, supporting joint evaluation of trajectory accuracy and downstream Gaussian reconstruction.}



\section{Method}
\label{sec:method}

\subsection{Overview}
\label{sec:method_overview}

\begin{figure}[t]
    \centering
    \includegraphics[width=\linewidth]{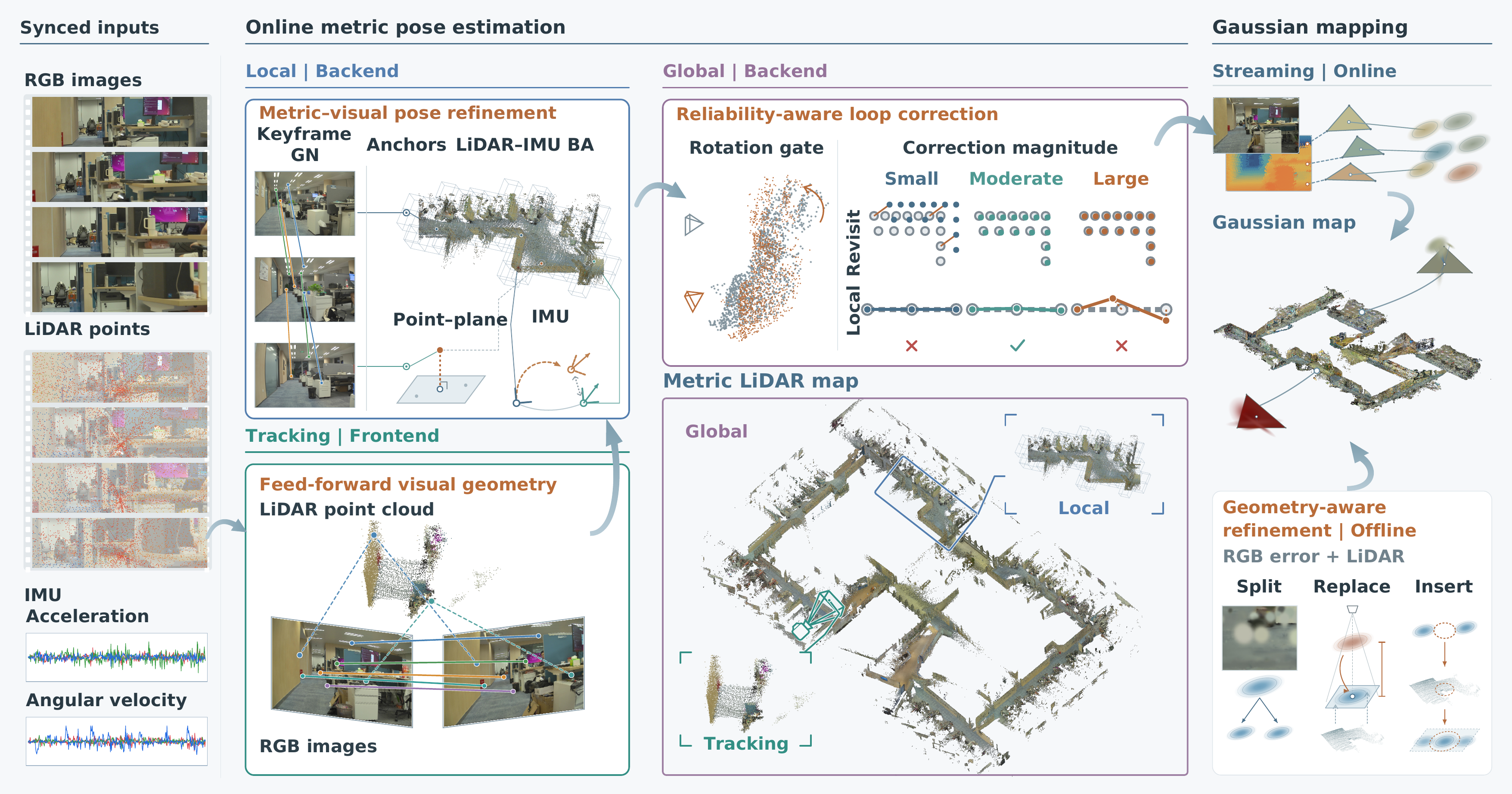}
    \caption{\textbf{Pipeline of MIRAGE.} We anchor feed-forward visual geometry to LiDAR--IMU motion for metric pose estimation and streaming Gaussian mapping. Local metric constraints support visual pose refinement, while reliability-aware loop correction controls correction magnitude and selectively applies rotation to balance revisit alignment and local motion consistency. The resulting poses drive a triangle-associated Gaussian map. With poses fixed, offline geometry-aware refinement improves scene structure using image residuals and LiDAR support.}
    \label{fig:mirage_pipeline}
\end{figure}

{Figure~\ref{fig:mirage_pipeline} summarizes MIRAGE. Given synchronized RGB, LiDAR, and IMU streams, we anchor feed-forward visual geometry to LiDAR--IMU motion for metric camera pose estimation. Correspondence support and cross-modal agreement regulate pose corrections (Section~\ref{sec:metric_pose}). The resulting poses drive streaming Gaussian reconstruction. After streaming, we fix camera poses and refine the Gaussian map offline, using image residuals and LiDAR depth to guide primitive splitting, replacement, and insertion (Section~\ref{sec:gaussian_refinement}). We also introduce a synchronized multimodal indoor benchmark with MIRAGE-derived 6-DoF reference trajectories for evaluating long-horizon monocular reconstruction (Section~\ref{sec:method_benchmark}).}

\subsection{Pose Estimation}
\label{sec:metric_pose}
{We anchor visual pose estimation to LiDAR--IMU motion through a local-to-global pipeline. \textbf{Tracking} combines dense visual correspondences with LiDAR-derived metric depth to estimate frame-to-keyframe poses (Section~\ref{sec:visual_pose_inference}). \textbf{Local optimization} refines keyframe poses using visual consistency constraints and metric position anchors (Section~\ref{sec:local_optimization}). \textbf{Loop closure} selectively corrects these anchors, and \textbf{global trajectory fusion} combines them with local visual motion to produce per-image camera poses (Section~\ref{sec:loop_correction}).}

\subsubsection{Metric-Guided Visual Tracking}
\label{sec:visual_pose_inference}
{For the current frame $f$ and reference keyframe $k$, we estimate the frame-to-keyframe transformation $\mathbf{T}_{kf}^{\mathrm{V}}=(\mathbf{T}_k^{\mathrm{V}})^{-1}\mathbf{T}_f^{\mathrm{V}}$, where $\mathbf{T}_f^{\mathrm{V}},\mathbf{T}_k^{\mathrm{V}}\in\mathrm{SE}(3)$ are their visual camera-to-world poses. As shown in Figure~\ref{fig:mirage_pipeline}, the feed-forward model~\citep{leroy2024grounding} predicts dense image correspondences and their confidence scores, while LiDAR projection and depth completion~\citep{wang2025depth} provide dense metric depth. We optimize $\mathbf{T}_{kf}^{\mathrm{V}}$ by minimizing a Huber loss over confidence-weighted pixel reprojection and log-depth residuals using Gauss--Newton, following~\citep{murai2025mast3rslam}. Details are provided in Appendix~\ref{app:visual_tracking}.}



\subsubsection{Local Optimization with LiDAR--Inertial Support}
\label{sec:local_optimization}
{After tracking and keyframe insertion, we couple visual keyframe optimization with LiDAR--inertial bundle adjustment through metric position anchors. Visual correspondences constrain inter-frame image and depth consistency, while LiDAR and IMU observations constrain metric motion. We first optimize the visual camera-to-world poses $\mathbf T_j^{\mathrm V}=(\mathbf R_j^{\mathrm V},\mathbf p_j^{\mathrm V})\in\mathrm{SE}(3)$ by minimizing}
\begin{equation}
E_{\mathrm V}
=
\sum_{(j,k)\in\mathcal E_{\mathrm V}}
\sum_{m\in\mathcal M_{jk}}
\rho\!\left(
\mathbf W_{jk,m}\,
\mathbf r^{\mathrm V}_{jk,m}
\!\left(
(\mathbf T_k^{\mathrm V})^{-1}\mathbf T_j^{\mathrm V}
\right)
\right).
\label{eq:local_visual_optimization}
\end{equation}
{Here, $\mathcal E_{\mathrm V}$ contains directed keyframe connections, and $\mathcal M_{jk}$ indexes valid correspondences. Using per-image metric depth, $\mathbf r^{\mathrm V}$ combines pixel reprojection and log-depth residuals; $\mathbf W$ encodes confidence and residual scaling, and $\rho$ sums componentwise Huber penalties. We fix the first keyframe pose to remove gauge freedom.}

{For LiDAR--inertial support, we organize multi-scan points into adaptive voxels and fit shared planes using PCA on locally planar clusters~\citep{liu2026voxelslam}. We separately refine motion within a sliding window by minimizing $E_{\mathrm{LI}}=E_{\mathrm{LiDAR}}+E_{\mathrm{IMU}}$. The LiDAR term penalizes point spread perpendicular to each plane, adjusting scan poses to reduce local point-cloud thickness and align observations across scans. The IMU term penalizes disagreement between estimated motion and integrated gyroscope and accelerometer measurements.}

{We interpolate the optimized LiDAR--inertial poses to keyframe timestamps, convert them to camera poses using calibrated extrinsics, and rigidly align them to the visual world frame. The resulting poses $\mathbf T_j^{\mathrm M}=(\mathbf R_j^{\mathrm M},\mathbf p_j^{\mathrm M})$ serve as metric anchors. We then refine visual keyframe positions by adding $E_{\mathrm{anchor}}=\sum_{j\in\mathcal S}\|\mathbf p_j^{\mathrm V}-\mathbf p_j^{\mathrm M}\|_2^2/\sigma_{\mathrm{BA}}^2$, where $\mathcal S$ indexes keyframes with available anchors and $\sigma_{\mathrm{BA}}$ controls the prior strength. Anchor positions remain fixed during refinement. Full pose anchors are retained for subsequent global correction and trajectory fusion (Section~\ref{sec:loop_correction}). Details are provided in Appendix~\ref{app:local_optimization}.}

\subsubsection{Loop Closure and Global Trajectory Fusion}
\label{sec:loop_correction}
{We selectively update the metric camera anchors through \textbf{observation-supported loop correction} and combine them with the locally refined visual trajectory through \textbf{metric-anchored visual motion fusion} to obtain per-image camera poses.}

\noindent\textbf{Observation-supported loop correction.}
{We generate initial loop candidates following~\citep{liu2026voxelslam} and rigidly align the resulting pose-graph optimization (PGO) proposals to the current metric trajectory without scale fitting. We assess each proposal from two complementary perspectives:
(1) \emph{geometric consistency}, evaluating improvements in revisit alignment and the risk of disrupting local structure; and
(2) \emph{visual consistency}, evaluating reprojection agreement with available image correspondences.
These cues determine the accepted correction magnitude. Rotation updates are gated separately and require additional geometric benefit relative to their risk; otherwise, the current metric orientations are retained, as shown in Figure~\ref{fig:mirage_pipeline}. Accepted corrections update the metric anchors and propagate to subsequent metric poses. Details are provided in Appendix~\ref{app:loop_correction}.}

\noindent\textbf{Metric-anchored visual motion fusion.}
{We interpolate consecutive metric anchors to obtain a baseline camera trajectory. We then incorporate visual motion between these anchors, using the deviation of the visual trajectory from the interpolation of its poses at the same two anchor timestamps. Let $\mathbf{T}^{\mathrm{M}}(t)$ interpolate the accepted metric anchors, $\mathbf{T}^{\mathrm{V}}(t)$ denote the locally refined visual trajectory, and $\overline{\mathbf{T}}^{\mathrm{V}}(t)$ interpolate the visual poses at the same two anchor timestamps. We obtain the fused camera pose $\mathbf{T}^{\mathrm{F}}(t)$ by combining their rotation and position components:}
\begin{equation}
\mathbf{T}^{\mathrm{F}}(t)
=
\left(
\mathbf{R}^{\mathrm{M}}(t)
\overline{\mathbf{R}}^{\mathrm{V}}(t)^\top
\mathbf{R}^{\mathrm{V}}(t),\;
\mathbf{p}^{\mathrm{M}}(t)
+
\mathbf{p}^{\mathrm{V}}(t)
-
\overline{\mathbf{p}}^{\mathrm{V}}(t)
\right).
\label{eq:global_visual_motion_fusion}
\end{equation}
{The fused trajectory matches the metric anchors exactly while retaining visual motion residuals between them. Details are provided in Appendix~\ref{app:trajectory_fusion}. We optionally refine the accepted metric trajectory offline using hierarchical LiDAR bundle adjustment~\citep{liu2023hba}, exporting a separate trajectory as described in Appendix~\ref{app:offline_hba}.}


\subsection{Geometry-Aware Gaussian Reconstruction}
\label{sec:gaussian_refinement}
{We reconstruct scenes using LiDAR to constrain geometry and RGB images to supervise appearance. Inspired by PLANING~\citep{jiang2026planing}, we use a triangle--Gaussian representation that couples explicit surface geometry with Gaussian appearance.}

\noindent\textbf{Streaming Gaussian reconstruction.} 
{For each incoming keyframe, we initialize triangle geometry from LiDAR depth and normals, and Gaussian appearance from RGB observations. We then jointly optimize geometry and appearance through differentiable rendering: depth and normal consistency constrain geometry, while photometric losses supervise appearance. When camera poses are updated, we transform the associated primitives accordingly to maintain map consistency. Details are provided in Appendix~\ref{app:gaussian_reconstruction}.}



\noindent\textbf{Geometry-aware Gaussian refinement.} 
{Following prior work~\citep{li2026artdeco,meuleman2025onthefly,zhang2025hi}, we refine the streamed scene offline to improve global scene consistency and reconstruction accuracy. With camera poses fixed, LiDAR depth constrains geometry, while RGB images refine appearance.}
{During refinement, we periodically diagnose and repair regions with high rendering error using motion-compensated LiDAR depth and multi-view consistency checks.
(1) Oversized Gaussians whose depths agree with LiDAR measurements but fail to resolve image detail are \textbf{split} into smaller primitives.
(2) Gaussians with persistent depth disagreement are \textbf{replaced} with primitives reinitialized using LiDAR depth and local normals.
(3) At locations with valid LiDAR depth but insufficient Gaussian coverage, we \textbf{insert} new primitives using depth for position, local normals for orientation, and RGB observations for color.
Alongside the photometric loss, we apply a point-to-plane penalty constructed from LiDAR depth and local normals to newly created or reinitialized Gaussians, penalizing position errors along the normal direction while allowing tangential adjustment. Repair criteria and loss details are provided in Appendix~\ref{app:gaussian_refinement}.}

\subsection{A Benchmark for Long-Horizon Indoor Reconstruction}
\label{sec:method_benchmark}

{We introduce a benchmark for long-horizon monocular streaming reconstruction. {Our platform (Fig.~\ref{fig:mirage_teaser}, right)} integrates a wide-angle RGB camera, two LiDARs, and one LiDAR's built-in IMU. The dual-LiDAR configuration combines broad coverage for robust localization with dense sampling for detailed reconstruction. Custom hardware synchronization and calibrated extrinsics align the sensor streams. We collected sequences through low-speed teleoperation to limit motion blur and removed blurred images through manual inspection.}

{The dataset comprises 10 sequences across four categories: laboratories and dining areas with numerous objects and reflective surfaces, an office, and long, low-texture corridors that challenge visual tracking. It contains 71,536 RGB images at $2880\times1860$ resolution, spanning 59.6 minutes and 1.15~km, with trajectories approaching 200~m. We provide synchronized RGB--LiDAR--IMU data, sensor intrinsics and extrinsics, global LiDAR reference point clouds, and MIRAGE-derived 6-DoF reference poses. MIRAGE's trajectory accuracy is validated on Oxford Spires~\citep{tao2026oxfordspires} and M2DGR~\citep{yin2022m2dgr}. Platform details and dataset statistics appear in Appendix~\ref{app:data_acquisition}.}

{We evaluate RGB-only methods on global trajectory accuracy and downstream reconstruction quality. We report absolute trajectory error (ATE) RMSE after Sim(3) alignment to the reference poses. For reconstruction evaluation, we train standard 3DGS~\citep{kerbl2023gaussian} with each method's predicted poses held fixed. Held-out novel view synthesis (NVS) PSNR serves as a downstream indicator of local multi-view pose consistency.}

\section{Experiments}
\label{sec:experiments}

\subsection{Experimental Setup}
\label{sec:experimental_setup}

\noindent\textbf{Datasets and Metrics.}
We evaluate MIRAGE on Oxford Spires~\citep{tao2026oxfordspires}, M2DGR~\citep{yin2022m2dgr}, and INS~\citep{deng2025mneslam}, reporting absolute trajectory error (ATE) RMSE for trajectory accuracy and PSNR, SSIM, and LPIPS~\citep{zhang2018unreasonable} for rendering quality. We additionally evaluate representative vision-only methods on our proposed MIRAGE benchmark, reporting ATE RMSE after Sim(3) alignment to the reference trajectories we provide. Details are provided in Appendix~\ref{app:implementation_evaluation}.

\noindent\textbf{Baselines.}
On Oxford Spires, M2DGR, and INS, our comparisons include LiDAR--inertial odometry (EllipseLIO~\citep{border2026ellipselio}), LiDAR--inertial--visual odometry (Fast-LIVO2~\citep{zheng2025fastlivo2}), LiDAR--inertial SLAM without visual input (Fast-LIO-SLAM and Voxel-SLAM~\citep{liu2026voxelslam}), and vision-only methods (LingBot-Map~\citep{chen2026lingbotmap}, GLUEMAP~\citep{pan2026gluemap}, and COLMAP~\citep{schoenberger2016sfm}). We also compare with Gaussian-LIC2~\citep{lang2026gaussianlic2}, a tightly coupled LiDAR--inertial--visual Gaussian SLAM system. On our MIRAGE benchmark, we evaluate recent strong vision-only methods, including LingBot-Map, GLUEMAP, LoGeR~\citep{zhang2026loger}, Scal3R~\citep{xie2026scal3r}, and ABot-Recon~\citep{han2026abotrecon}, alongside MASt3R-SLAM~\citep{murai2025mast3rslam}.

\noindent\textbf{Implementation Details.}
Following the 3DGS evaluation protocol~\citep{kerbl2023gaussian}, we hold out every eighth image for rendering evaluation. LiDAR--inertial local bundle adjustment uses a ten-scan sliding window. Gaussian refinement runs for 15,000 steps with camera poses and exposure transforms fixed, performing splitting every 1,000 steps from step 2,000 through 8,000. All experiments are conducted on NVIDIA RTX 4090 GPUs.

\subsection{Trajectory Accuracy}
\label{sec:experiments_pose}
{MIRAGE completes all 15 sequences and achieves the lowest reported online ATE on 12 while maintaining a moderate latency of 0.73~s/frame (Table~\ref{tab:main_pose_results}). Averaged equally across sequences, its ATE is 4.22~cm versus 7.49~cm for EllipseLIO, a 43.7\% reduction. On large-scale Oxford Spires traversals and low-texture INS scenes, LingBot-Map exhibits substantial trajectory errors, while Fast-LIVO2 and Gaussian-LIC2 show severe errors or failures in our reproductions. MIRAGE maintains ATE below 10~cm on every sequence. These results are consistent with its reliability-aware design: metric anchors constrain drift, while correspondence support and cross-modal agreement regulate visual corrections to exploit useful evidence and limit unreliable updates. With offline HBA, MIRAGE achieves a mean ATE of 2.96~cm versus 3.31~cm for Voxel-SLAM, with lower errors on 11 of 15 sequences.}

\begin{table}[t]
\centering
\caption{Translational ATE RMSE (cm, $\downarrow$). Vision-only and metric methods use Sim(3) and SE(3) alignment, respectively. Red/orange mark the best/second-best per sequence, ranked separately for online and offline methods; -- denotes failure.}
\label{tab:main_pose_results}
\begingroup
\setlength{\tabcolsep}{2.5pt}
\renewcommand{\arraystretch}{1.18}
\scriptsize
\resizebox{\linewidth}{!}{%
\begin{tabular}{lrrrrrr|rrrr}
\toprule
& \multicolumn{6}{c|}{Online} & \multicolumn{4}{c}{Offline} \\
\cmidrule(lr){2-7}\cmidrule(lr){8-11}
Sequence &
EllipseLIO &
\shortstack{Fast-LIO-\\SLAM} &
\shortstack{Fast-\\LIVO2} &
\shortstack{Gaussian-\\LIC2} &
\shortstack{LingBot-\\Map} &
\textbf{Ours} &
\shortstack{Voxel-\\SLAM} &
COLMAP &
GLUEMAP &
\textbf{Ours+HBA} \\
Latency &
$0.10$ &
$0.12$ &
$0.10$ &
$0.71$ &
$0.18$ &
$0.73$ &
$0.145$ &
$1.85$ &
$10.65$ &
$0.76$ \\
\midrule
\multicolumn{11}{l}{\textit{Oxford Spires}} \\
Observatory~01 & $5.74$ & $6.28$ & \cellcolor{orange!20}$3.97$ & $6.41\!\times\!10^{4}$ & $2.46\!\times\!10^{2}$ & \cellcolor{red!35}$3.43$ & \cellcolor{orange!20}$2.46$ & $6.66$ & $95.79$ & \cellcolor{red!35}$2.09$ \\
Observatory~02 & \cellcolor{orange!20}$5.23$ & $26.22$ & $6.83$ & -- & $3.57\!\times\!10^{2}$ & \cellcolor{red!35}$3.56$ & \cellcolor{orange!20}$2.14$ & $7.69$ & $75.81$ & \cellcolor{red!35}$2.07$ \\
Keble~02 & \cellcolor{orange!20}$6.02$ & $37.35$ & $3.01\!\times\!10^{5}$ & $3.08\!\times\!10^{2}$ & $3.27\!\times\!10^{2}$ & \cellcolor{red!35}$3.60$ & \cellcolor{red!35}$3.48$ & $4.74$ & $3.30\!\times\!10^{3}$ & \cellcolor{orange!20}$3.66$ \\
Keble~03 & \cellcolor{red!35}$6.02$ & $33.26$ & $7.92$ & $2.32\!\times\!10^{8}$ & $3.29\!\times\!10^{2}$ & \cellcolor{orange!20}$7.13$ & \cellcolor{orange!20}$3.60$ & $4.51$ & $3.09\!\times\!10^{3}$ & \cellcolor{red!35}$3.07$ \\
Keble~04 & $19.14$ & $46.38$ & \cellcolor{red!35}$8.82$ & $1.50\!\times\!10^{5}$ & $3.19\!\times\!10^{2}$ & \cellcolor{orange!20}$9.61$ & $8.37$ & \cellcolor{orange!20}$7.03$ & $1.14\!\times\!10^{2}$ & \cellcolor{red!35}$6.52$ \\
Blenheim~01 & $23.58$ & $58.42$ & \cellcolor{orange!20}$14.04$ & $1.17\!\times\!10^{4}$ & $4.12\!\times\!10^{2}$ & \cellcolor{red!35}$4.50$ & \cellcolor{orange!20}$3.88$ & $7.95$ & $5.65\!\times\!10^{3}$ & \cellcolor{red!35}$3.52$ \\
Blenheim~02 & \cellcolor{orange!20}$11.20$ & $11.30$ & $23.18$ & $1.99\!\times\!10^{5}$ & $3.77\!\times\!10^{2}$ & \cellcolor{red!35}$3.27$ & \cellcolor{orange!20}$2.50$ & $5.50$ & $3.83\!\times\!10^{3}$ & \cellcolor{red!35}$2.31$ \\
Church~03 & \cellcolor{orange!20}$2.05$ & $29.44$ & $2.73\!\times\!10^{6}$ & -- & $1.36\!\times\!10^{2}$ & \cellcolor{red!35}$1.15$ & \cellcolor{red!35}$1.35$ & $6.86$ & $3.77\!\times\!10^{2}$ & \cellcolor{orange!20}$1.50$ \\
\midrule
\multicolumn{11}{l}{\textit{M2DGR}} \\
room\_01 & $2.87$ & $24.57$ & \cellcolor{orange!20}$2.59$ & $2.97$ & $13.12$ & \cellcolor{red!35}$1.87$ & \cellcolor{orange!20}$1.72$ & $31.26$ & $23.84$ & \cellcolor{red!35}$1.67$ \\
room\_02 & $2.65$ & $19.27$ & \cellcolor{orange!20}$2.40$ & $2.64$ & $25.75$ & \cellcolor{red!35}$2.18$ & \cellcolor{red!35}$1.89$ & $11.56$ & $27.71$ & \cellcolor{orange!20}$1.90$ \\
room\_03 & \cellcolor{orange!20}$2.43$ & $24.62$ & $2.91$ & $3.86$ & $16.48$ & \cellcolor{red!35}$2.15$ & \cellcolor{orange!20}$2.10$ & $15.69$ & $15.55$ & \cellcolor{red!35}$2.00$ \\
\midrule
\multicolumn{11}{l}{\textit{INS}} \\
indoor\_sq1 & $10.21$ & \cellcolor{orange!20}$9.96$ & $11.22$ & $15.37$ & $2.06\!\times\!10^{2}$ & \cellcolor{red!35}$5.67$ & \cellcolor{orange!20}$4.02$ & $52.99$ & $1.95\!\times\!10^{3}$ & \cellcolor{red!35}$2.96$ \\
indoor\_sq2 & \cellcolor{red!35}$4.35$ & $10.19$ & $15.16$ & -- & $1.45\!\times\!10^{2}$ & \cellcolor{orange!20}$5.48$ & \cellcolor{orange!20}$4.38$ & -- & $1.93\!\times\!10^{3}$ & \cellcolor{red!35}$3.37$ \\
indoor\_sq3 & \cellcolor{orange!20}$6.03$ & $6.50$ & $10.52$ & $11.62$ & $2.44\!\times\!10^{2}$ & \cellcolor{red!35}$5.20$ & \cellcolor{orange!20}$3.92$ & $52.33$ & $1.88\!\times\!10^{3}$ & \cellcolor{red!35}$3.70$ \\
indoor\_sq4 & \cellcolor{orange!20}$4.77$ & $8.30$ & $14.20$ & $12.28$ & $3.11\!\times\!10^{2}$ & \cellcolor{red!35}$4.44$ & \cellcolor{red!35}$3.90$ & $129.19$ & $2.56\!\times\!10^{2}$ & \cellcolor{orange!20}$4.03$ \\
\bottomrule
\end{tabular}%
}
\endgroup

\end{table}

\subsection{Rendering Quality}
\label{sec:experiments_rendering}
\begin{figure}[t]
    \centering
    \includegraphics[width=\linewidth]{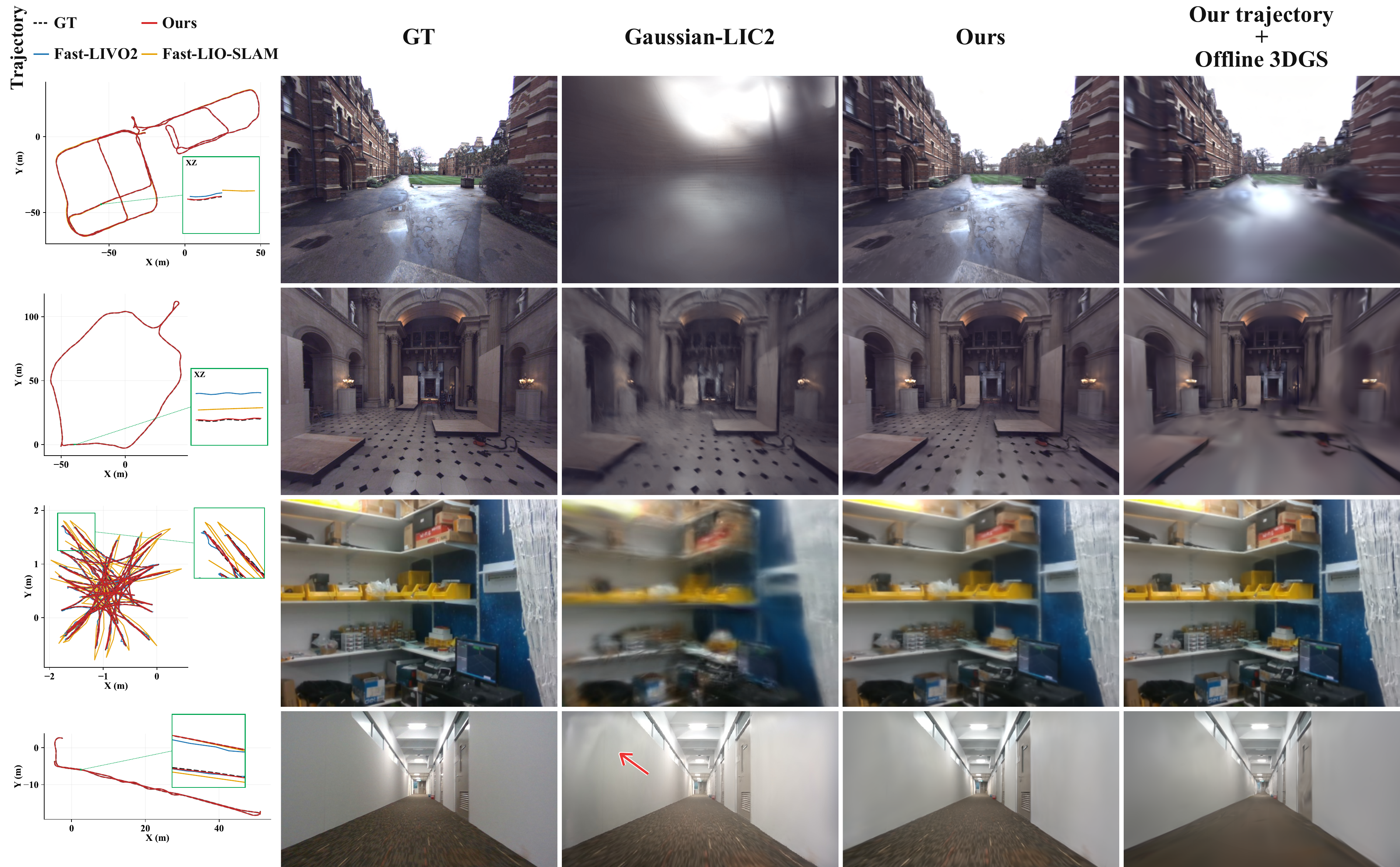}
    \caption{\textbf{Qualitative trajectory and rendering comparisons.} Left: SE(3)-aligned XY trajectories; green boxes show XZ close-ups with equal axis scales. Right: rendering comparisons.}
    \label{fig:qualitative_comparison}
\end{figure}

{We compare MIRAGE with Gaussian-LIC2 and offline 3DGS trained from scratch using MIRAGE trajectories (Table~\ref{tab:main_rendering_results}). We evaluate 12 sequences, excluding three for which Gaussian-LIC2 did not complete reconstruction. MIRAGE achieves higher PSNR than Gaussian-LIC2 on 10 of 12 sequences, with a mean gain of 3.27~dB. Using identical camera trajectories, our refined streaming maps also exceed offline 3DGS by 1.15~dB in mean PSNR. These results are consistent with the complementary benefits of reliable multi-view alignment and Gaussian refinement that combines LiDAR-guided structural repair with RGB appearance supervision to recover scene detail. Figure~\ref{fig:qualitative_comparison} presents representative trajectory and rendering comparisons across the three datasets.}


\begin{table}[t]
\centering
\caption{Held-out rendering quality. Offline 3DGS~\citep{kerbl2023gaussian} uses 30k iterations with MIRAGE's online poses fixed. Cells list PSNR (dB, $\uparrow$), SSIM ($\uparrow$), and LPIPS ($\downarrow$) from top to bottom. Red and orange indicate the best and second-best values per sequence and metric.}
\label{tab:main_rendering_results}
\begingroup
\scriptsize
\setlength{\fboxsep}{1.2pt}
\newcommand{\renderrank}[2]{\colorbox{#1}{\makebox[3.2em][r]{\strut #2}}}
\setlength{\tabcolsep}{2.5pt}
\renewcommand{\arraystretch}{1.12}
\resizebox{\linewidth}{!}{%
\begin{tabular}{lrrrrrrrrrrrr}
\toprule
& \multicolumn{6}{c}{\textit{Oxford Spires}} & \multicolumn{3}{c}{\textit{M2DGR}} & \multicolumn{3}{c}{\textit{INS}} \\
\cmidrule(lr){2-7}\cmidrule(lr){8-10}\cmidrule(lr){11-13}
Method & \makecell[c]{Observatory\\01} & \makecell[c]{Keble\\02} & \makecell[c]{Keble\\03} & \makecell[c]{Keble\\04} & \makecell[c]{Blenheim\\01} & \makecell[c]{Blenheim\\02} & room\_01 & room\_02 & room\_03 & indoor\_sq1 & indoor\_sq3 & indoor\_sq4 \\
\midrule
Gaussian-LIC2 & \makecell[r]{\renderrank{white}{17.91}\\\renderrank{white}{0.678}\\\renderrank{white}{0.570}} & \makecell[r]{\renderrank{white}{24.26}\\\renderrank{white}{0.810}\\\renderrank{white}{0.421}} & \makecell[r]{\renderrank{white}{17.49}\\\renderrank{white}{0.678}\\\renderrank{white}{0.554}} & \makecell[r]{\renderrank{white}{18.72}\\\renderrank{white}{0.680}\\\renderrank{white}{0.548}} & \makecell[r]{\renderrank{white}{21.76}\\\renderrank{white}{0.738}\\\renderrank{white}{0.530}} & \makecell[r]{\renderrank{white}{21.28}\\\renderrank{white}{0.744}\\\renderrank{white}{0.535}} & \makecell[r]{\renderrank{white}{25.15}\\\renderrank{white}{0.809}\\\renderrank{white}{0.258}} & \makecell[r]{\renderrank{white}{26.67}\\\renderrank{white}{0.829}\\\renderrank{orange!20}{0.251}} & \makecell[r]{\renderrank{white}{24.60}\\\renderrank{white}{0.770}\\\renderrank{white}{0.341}} & \makecell[r]{\renderrank{red!35}{30.09}\\\renderrank{orange!20}{0.893}\\\renderrank{orange!20}{0.299}} & \makecell[r]{\renderrank{orange!20}{29.24}\\\renderrank{orange!20}{0.885}\\\renderrank{orange!20}{0.308}} & \makecell[r]{\renderrank{red!35}{30.45}\\\renderrank{orange!20}{0.896}\\\renderrank{red!35}{0.301}} \\
\addlinespace[3pt]
\textbf{Ours} & \makecell[r]{\renderrank{red!35}{26.26}\\\renderrank{red!35}{0.827}\\\renderrank{red!35}{0.308}} & \makecell[r]{\renderrank{red!35}{27.53}\\\renderrank{red!35}{0.836}\\\renderrank{red!35}{0.325}} & \makecell[r]{\renderrank{red!35}{26.80}\\\renderrank{red!35}{0.821}\\\renderrank{red!35}{0.344}} & \makecell[r]{\renderrank{red!35}{24.33}\\\renderrank{red!35}{0.741}\\\renderrank{red!35}{0.390}} & \makecell[r]{\renderrank{red!35}{26.40}\\\renderrank{red!35}{0.796}\\\renderrank{red!35}{0.396}} & \makecell[r]{\renderrank{red!35}{27.82}\\\renderrank{red!35}{0.837}\\\renderrank{red!35}{0.357}} & \makecell[r]{\renderrank{orange!20}{25.44}\\\renderrank{orange!20}{0.824}\\\renderrank{orange!20}{0.255}} & \makecell[r]{\renderrank{orange!20}{26.76}\\\renderrank{orange!20}{0.831}\\\renderrank{white}{0.252}} & \makecell[r]{\renderrank{red!35}{26.60}\\\renderrank{orange!20}{0.843}\\\renderrank{orange!20}{0.249}} & \makecell[r]{\renderrank{orange!20}{30.03}\\\renderrank{red!35}{0.907}\\\renderrank{red!35}{0.294}} & \makecell[r]{\renderrank{red!35}{29.58}\\\renderrank{red!35}{0.901}\\\renderrank{red!35}{0.306}} & \makecell[r]{\renderrank{orange!20}{29.34}\\\renderrank{red!35}{0.900}\\\renderrank{orange!20}{0.307}} \\
\addlinespace[3pt]
\makecell[l]{\textbf{Our trajectory} +\\Offline 3DGS} & \makecell[r]{\renderrank{orange!20}{26.04}\\\renderrank{orange!20}{0.801}\\\renderrank{orange!20}{0.389}} & \makecell[r]{\renderrank{orange!20}{26.87}\\\renderrank{orange!20}{0.823}\\\renderrank{orange!20}{0.366}} & \makecell[r]{\renderrank{orange!20}{25.75}\\\renderrank{orange!20}{0.804}\\\renderrank{orange!20}{0.400}} & \makecell[r]{\renderrank{orange!20}{23.19}\\\renderrank{orange!20}{0.711}\\\renderrank{orange!20}{0.476}} & \makecell[r]{\renderrank{orange!20}{24.82}\\\renderrank{orange!20}{0.749}\\\renderrank{orange!20}{0.507}} & \makecell[r]{\renderrank{orange!20}{26.65}\\\renderrank{orange!20}{0.787}\\\renderrank{orange!20}{0.470}} & \makecell[r]{\renderrank{red!35}{26.69}\\\renderrank{red!35}{0.865}\\\renderrank{red!35}{0.212}} & \makecell[r]{\renderrank{red!35}{27.02}\\\renderrank{red!35}{0.847}\\\renderrank{red!35}{0.231}} & \makecell[r]{\renderrank{orange!20}{26.42}\\\renderrank{red!35}{0.850}\\\renderrank{red!35}{0.248}} & \makecell[r]{\renderrank{white}{26.36}\\\renderrank{white}{0.831}\\\renderrank{white}{0.451}} & \makecell[r]{\renderrank{white}{27.71}\\\renderrank{white}{0.855}\\\renderrank{white}{0.414}} & \makecell[r]{\renderrank{white}{25.60}\\\renderrank{white}{0.832}\\\renderrank{white}{0.453}} \\
\bottomrule
\end{tabular}%
}
\endgroup
\end{table}

\subsection{Ablation Studies}
\label{sec:experiments_ablation}

\noindent\textbf{Tracking, Local Optimization, and Global Correction.}
{Table~\ref{tab:ablation_summary}(a) shows that local optimization substantially reduces mean ATE across datasets, while global correction provides further gains with occasional small regressions. Oxford Spires shows the largest overall improvement, with mean ATE decreasing from 82.36~cm after tracking to 4.53~cm after global correction. Offline HBA results are included as a separate reference.}



\noindent\textbf{Visual Evidence and Gaussian Refinement.}
{Table~\ref{tab:ablation_summary}(b) shows that removing the visual pose and geometry branch substantially degrades rendering, highlighting the joint contribution of visual tracking and dense geometric priors. Both refinement variants improve upon the full streaming model. Under the same 15k-step budget with poses and exposure fixed, geometry-aware refinement exceeds photometric refinement by 0.44~dB in mean PSNR, with higher SSIM and lower LPIPS on all three sequences.}

\begin{table}[t]
\centering
\caption{Pose and rendering ablations. (a) Scene-averaged ATE (cm, $\downarrow$) over all evaluated sequences. (b) PSNR (dB, $\uparrow$) / SSIM ($\uparrow$) / LPIPS ($\downarrow$). $\dagger$: camera poses are obtained solely by interpolating LiDAR--inertial poses and applying calibrated extrinsics; RGB supervision for Gaussian reconstruction is retained.}
\label{tab:ablation_summary}
\label{tab:pose_stages}
\label{tab:rendering_ablation}
\begingroup
\footnotesize
\setlength{\tabcolsep}{2pt}
\renewcommand{\arraystretch}{1.12}
\resizebox{\linewidth}{!}{%
\begin{tabular}{@{}lccc|lccc@{}}
\toprule
\multicolumn{4}{c|}{\textbf{(a) Pose estimation}} & \multicolumn{4}{c}{\textbf{(b) Rendering on M2DGR}} \\
\cmidrule(lr){1-4}\cmidrule(lr){5-8}
Stage & \makecell[c]{Oxford\\Spires} & M2DGR & INS & Configuration & room\_01 & room\_02 & room\_03 \\
\midrule
Tracking & 82.363 & 21.868 & 43.819 & w/o visual$^\dagger$ & 13.90 / 0.504 / 0.561 & 13.04 / 0.512 / 0.559 & 12.89 / 0.489 / 0.577 \\
Local & 7.784 & 1.909 & 7.055 & Streaming & 23.28 / 0.746 / 0.339 & 24.41 / 0.762 / 0.344 & 24.74 / 0.784 / 0.319 \\
Global & 4.529 & 2.067 & 5.195 & + Photo. refine & 24.96 / 0.811 / 0.272 & 26.29 / 0.816 / 0.275 & 26.22 / 0.832 / 0.265 \\
Offline (+HBA) & 3.093 & 1.858 & 3.513 & + Geo. refine & 25.44 / 0.824 / 0.255 & 26.76 / 0.831 / 0.252 & 26.60 / 0.843 / 0.249 \\
\bottomrule
\end{tabular}%
}
\endgroup
\end{table}

\subsection{Vision-Only Evaluation on the MIRAGE Benchmark}
\label{sec:experiments_mirage_benchmark}

{Table~\ref{tab:mirage_visual_pose_results} evaluates RGB-only pose estimates through global trajectory accuracy and downstream fixed-pose 3DGS rendering. ABot-Recon achieves the lowest mean Sim(3)-aligned ATE among methods completing all ten sequences (1.58~m), while GLUEMAP achieves the highest mean downstream PSNR among visual baselines (22.80~dB). Their differing rankings suggest that global trajectory agreement alone does not determine reconstruction quality: local multi-view pose consistency also matters. Long indoor traversals, repetitive structures, and low-texture corridors challenge drift control and correspondence estimation, exposing complementary strengths and limitations of current visual methods.}

\begin{table}[t]
\centering
\caption{MIRAGE benchmark. ATE: Sim(3)-aligned translational RMSE against MIRAGE references. Offline 3DGS: held-out PSNR with fixed poses; Ours (reference) uses reference poses. Stream PSNR: MIRAGE after refinement. Red/orange indicate the best/second-best per sequence within the ATE and offline 3DGS groups; -- denotes trajectory failure.}
\label{tab:mirage_visual_pose_results}
\begingroup
\small
\setlength{\tabcolsep}{3pt}
\renewcommand{\arraystretch}{1.15}
\resizebox{\linewidth}{!}{%
\begin{tabular}{lc|cccccc|ccccccc|c}
\toprule
& & \multicolumn{6}{c|}{ATE (m, $\downarrow$)} & \multicolumn{7}{c|}{Offline 3DGS PSNR (dB, $\uparrow$)} & \shortstack{Stream\\PSNR} \\
\cmidrule(lr){3-8}\cmidrule(lr){9-15}\cmidrule(lr){16-16}
Sequence & \shortstack{Trajectory\\length (m)} & \shortstack{LingBot-\\Map} & GlueMap & LoGeR & Scal3R & \shortstack{ABot-\\Recon} & \shortstack{MASt3R-\\SLAM} & \shortstack{LingBot-\\Map} & GlueMap & LoGeR & Scal3R & \shortstack{ABot-\\Recon} & \shortstack{MASt3R-\\SLAM} & \shortstack{Ours\\(reference)} & Ours \\
\midrule
Laboratory 1 & 54.57 & \cellcolor{orange!20}0.299 & \cellcolor{red!35}0.205 & 0.927 & 1.901 & 0.368 & 0.491 & 21.65 & \cellcolor{red!35}26.59 & 20.74 & 22.16 & 23.28 & 17.54 & \cellcolor{orange!20}26.10 & 27.47 \\
Laboratory 2 & 61.02 & \cellcolor{orange!20}0.430 & \cellcolor{red!35}0.225 & 0.802 & 1.102 & 0.675 & 0.866 & 19.93 & \cellcolor{red!35}26.41 & 19.94 & 22.11 & 22.53 & 14.59 & \cellcolor{orange!20}25.02 & 28.47 \\
Laboratory 3 & 168.46 & 9.523 & 15.851 & 7.364 & 1.662 & \cellcolor{orange!20}0.930 & \cellcolor{red!35}0.729 & 18.51 & \cellcolor{orange!20}22.60 & 18.96 & 21.72 & 22.06 & 17.22 & \cellcolor{red!35}23.56 & 26.74 \\
\midrule
Dining Area 1 & 112.69 & 5.058 & 15.723 & 10.616 & 14.772 & \cellcolor{red!35}0.903 & \cellcolor{orange!20}2.168 & 18.42 & \cellcolor{red!35}22.16 & 18.76 & 19.09 & 21.33 & 16.16 & \cellcolor{orange!20}21.68 & 26.15 \\
Dining Area 2 & 64.65 & \cellcolor{orange!20}0.316 & \cellcolor{red!35}0.206 & 1.511 & 0.350 & 0.427 & 0.412 & 20.11 & 20.45 & 19.45 & \cellcolor{orange!20}20.50 & \cellcolor{red!35}20.87 & 16.41 & 19.84 & 24.06 \\
Dining Area 3 & 181.52 & 8.436 & \cellcolor{red!35}0.746 & 6.447 & 10.129 & \cellcolor{orange!20}1.656 & 3.836 & 17.40 & \cellcolor{orange!20}22.35 & 18.34 & 20.87 & 21.65 & 15.34 & \cellcolor{red!35}22.91 & 26.42 \\
Dining Area 4 & 65.37 & 0.545 & \cellcolor{red!35}0.149 & 0.376 & 0.602 & \cellcolor{orange!20}0.361 & 0.790 & 21.40 & \cellcolor{red!35}25.72 & 21.43 & 24.33 & 24.10 & 16.56 & \cellcolor{orange!20}24.69 & 26.72 \\
\midrule
Corridor 1 & 165.39 & 14.914 & 13.560 & 13.153 & \cellcolor{orange!20}6.802 & \cellcolor{red!35}1.873 & -- & 20.39 & 23.47 & 21.01 & 23.76 & \cellcolor{orange!20}23.86 & -- & \cellcolor{red!35}25.41 & 29.12 \\
Corridor 2 & 85.57 & \cellcolor{orange!20}3.647 & 11.387 & 7.694 & 9.206 & \cellcolor{red!35}1.026 & -- & 21.60 & 22.29 & 23.02 & 25.83 & \cellcolor{orange!20}26.24 & -- & \cellcolor{red!35}27.87 & 30.89 \\
\midrule
Office 1 & 195.15 & 21.861 & 23.101 & \cellcolor{orange!20}16.040 & 21.136 & \cellcolor{red!35}7.600 & -- & 15.15 & 15.92 & 15.68 & 16.25 & \cellcolor{orange!20}17.80 & -- & \cellcolor{red!35}18.57 & 24.35 \\
\bottomrule
\end{tabular}%
}
\endgroup
\end{table}

\section{Conclusion}
{We presented MIRAGE, which anchors feed-forward visual priors to LiDAR--IMU motion for metric pose estimation and Gaussian reconstruction. Visual and geometric evidence guide trajectory corrections, while LiDAR depth and image residuals drive Gaussian repair with camera poses fixed. Experiments on Oxford Spires, M2DGR, and INS demonstrate accurate trajectories and high-fidelity rendering. Our synchronized multimodal benchmark comprises 10 sequences across four indoor scene categories, with 71,536 RGB images spanning 59.6 minutes and 1.15~km. With MIRAGE-derived 6-DoF reference trajectories, sensor calibration, and global LiDAR maps, the benchmark provides the community with a common testbed for evaluating trajectory accuracy and downstream novel-view synthesis in monocular streaming reconstruction, supporting further advances in long-horizon indoor reconstruction.}

\section{AI use statement}
In this work, we used generative AI tools solely to assist with language polishing of the manuscript, including improving grammar, clarity, and readability. We did not use generative AI tools to generate research ideas, develop methods, analyze data, or produce experimental results. We have reviewed all AI-assisted edits to ensure that they preserve the intended meaning and technical accuracy of the manuscript and do not introduce unsupported claims. We take responsibility for the final content of this work, including text, claims, or artifacts produced with the aid of generative AI.

\bibliography{iclr2027_conference}
\bibliographystyle{iclr2027_conference}

\appendix
\section{Appendix}
\label{sec:appendix}

This appendix is organized as follows:
\begin{itemize}[leftmargin=1.2em]
    \item Section~\ref{app:notation} defines notation and coordinate conventions.
    \item Section~\ref{app:pose_estimation} details visual tracking, local optimization, loop correction, trajectory fusion, and optional offline pose refinement.
    \item Section~\ref{app:gaussian_reconstruction} details streaming Gaussian reconstruction and geometry-aware refinement.
    \item Section~\ref{app:data_acquisition} describes the acquisition platform, calibration, synchronization, collection protocol, and dataset statistics.
    \item Section~\ref{app:implementation_evaluation} specifies implementation settings and evaluation protocols.
    \item Section~\ref{app:additional_results} presents additional experimental results.
    \item Section~\ref{app:limitations} discusses limitations and future directions.
\end{itemize}

\subsection{Notation and Coordinate Conventions}
\label{app:notation}
{We use bold upright symbols for vectors and matrices. A rigid transformation $\mathbf{T}_{AB}=(\mathbf{R}_{AB},\mathbf{p}_{AB})\in\mathrm{SE}(3)$ maps coordinates from frame $B$ to frame $A$ as $\mathbf{x}_A=\mathbf{R}_{AB}\mathbf{x}_B+\mathbf{p}_{AB}$. Camera poses follow the camera-to-world convention, so their translations represent camera centers in world coordinates. Table~\ref{tab:notation} summarizes the notation used throughout.}

\begin{table}[!t]
    \centering
    \caption{Notation and coordinate conventions.}
    \label{tab:notation}
    \small
    \renewcommand{\arraystretch}{1.15}
    \begin{tabular}{@{}p{0.30\linewidth}p{0.64\linewidth}@{}}
        \toprule
        Symbol & Definition \\
        \midrule
        $\mathrm{W},\mathrm{C},\mathrm{L},\mathrm{B}$
        & World, camera, LiDAR, and IMU coordinate frames. \\

        $\mathbf{T}^{\mathrm{V}},\mathbf{T}^{\mathrm{M}},\mathbf{T}^{\mathrm{F}}$
        & Visual camera poses, metric camera-pose anchors, and fused camera poses. \\

        $\mathbf{R}\in\mathrm{SO}(3),\ \mathbf{p}\in\mathbb{R}^3$
        & Rotation and translation components of a pose. \\

        $f;\ j,k$
        & Current image index; visual keyframe indices. \\

        $i;\ m$
        & LiDAR--IMU temporal index; image correspondence index. \\

        $\tau_i,\ a_i$
        & Sensor acquisition time and corresponding pose-estimate availability time, with $a_i\geq\tau_i$. \\

        $\mathbf{K}$
        & Calibrated camera intrinsic matrix. \\

        $D_f(\mathbf{u})$
        & Depth at pixel $\mathbf{u}$ in image $f$. \\
        \bottomrule
    \end{tabular}
\end{table}

{Temporal interpolation uses acquisition timestamps $\tau_i$; estimates are consumed only after their availability times $a_i$. Depth $D_f(\mathbf{u})$ denotes the camera-frame $z$ coordinate of the corresponding 3D point, rather than its Euclidean distance to the camera center. Additional symbols are defined when first introduced.}

\subsection{Pose Estimation Details}
\label{app:pose_estimation}

\subsubsection{Metric-Guided Visual Tracking}
\label{app:visual_tracking}

{We use EllipseLIO~\citep{border2026ellipselio} to estimate the LiDAR--IMU trajectory for motion compensation. Using this trajectory, we align points from five recent LiDAR scans to the camera frame at the image timestamp and project them to obtain sparse metric depth. Depth completion~\citep{wang2025depth} converts the sparse projections into dense depth; invalid pixels are filled from visual depth predictions scaled using valid metric overlap. The feed-forward model~\citep{leroy2024grounding} supplies image correspondences $(\mathbf{u}_m,\mathbf{v}_m)$ and their confidence scores. The valid set $\mathcal M_{fk}$ retains matches passing geometric, confidence, positive-depth, and projection-validity checks.}

{To optimize the frame-to-keyframe pose $\mathbf{T}_{kf}^{\mathrm{V}}$ using these correspondences and metric depths, we initialize translation from LiDAR--IMU motion expressed in camera coordinates and retain the visual rotation seed. We then jointly refine rotation and translation with the reference keyframe fixed:}
\begin{equation}
\begin{aligned}
\mathbf{T}_{kf}^{\mathrm{V}}
&=
\arg\min_{\mathbf{T}\in\mathrm{SE}(3)}
\sum_{m\in\mathcal M_{fk}}
\rho\!\left(
\mathbf{W}_{fk,m}\,
\mathbf{r}^{\mathrm{V}}_{fk,m}(\mathbf{T})
\right),\\
\mathbf{r}^{\mathrm{V}}_{fk,m}(\mathbf{T})
&=
\begin{bmatrix}
\pi_{\mathbf{K}}\!\left(
\mathbf{T}\mathbf{X}_f(\mathbf{u}_m)
\right)-\mathbf{v}_m\\[2pt]
\displaystyle
\log\!\frac{
[\mathbf{T}\mathbf{X}_f(\mathbf{u}_m)]_z
}{
D_k(\mathbf{v}_m)
}
\end{bmatrix},\\
\mathbf{W}_{fk,m}
&=
\sqrt{q_{fk,m}}\,
\operatorname{diag}\!\left(
\sigma_{\mathrm{pix}}^{-1},
\sigma_{\mathrm{pix}}^{-1},
\sigma_{\log d}^{-1}
\right).
\end{aligned}
\label{eq:metric_visual_tracking}
\end{equation}
{Here, $\mathbf{X}_f(\mathbf{u}_m)=D_f(\mathbf{u}_m)\mathbf{K}^{-1}\widetilde{\mathbf{u}}_m$ is the back-projected 3D point, with $\widetilde{\mathbf{u}}_m$ denoting homogeneous pixel coordinates. The operator $\pi_{\mathbf{K}}$ performs perspective projection, $[\cdot]_z$ extracts axial depth, and $q_{fk,m}$ is the geometric mean of the matched confidence scores. The scales $\sigma_{\mathrm{pix}}$ and $\sigma_{\log d}$ normalize pixel reprojection and log-depth residuals, respectively, while $\rho$ sums componentwise Huber penalties. Following~\citep{murai2025mast3rslam}, we use Gauss--Newton with Huber reweighting and left updates $\mathbf{T}\leftarrow\exp(\mathbf{d}^{\wedge})\mathbf{T}$, where $\mathbf{d}\in\mathbb{R}^6$ is the pose increment and $(\cdot)^\wedge$ maps it to $\mathfrak{se}(3)$. Depths and correspondences remain fixed during each pose solve.}

\subsubsection{Local Optimization with LiDAR--Inertial Support}
\label{app:local_optimization}

\noindent\textbf{Visual keyframe optimization.}
{We optimize visual keyframes using the bidirectional pixel reprojection and log-depth factors in Equation~\ref{eq:local_visual_optimization}, keeping metric depths fixed and fixing the first keyframe pose to remove gauge freedom.}

\noindent\textbf{LiDAR--inertial bundle adjustment.}
{Following~\citep{liu2026voxelslam}, we separately optimize a sliding window of $N$ LiDAR--inertial poses, velocities, and biases:}
\begin{equation}
\begin{aligned}
\mathcal X_{\mathcal W}
&=
\left\{
\left(
\mathbf T_i^{\mathrm{LI}},
\mathbf v_i,
\mathbf b_i^{g},
\mathbf b_i^{a}
\right)
\right\}_{i=1}^{N},\\
\mathcal X_{\mathcal W}^{\star}
&=
\arg\min_{\mathcal X_{\mathcal W}}
E_{\mathrm{LiDAR}}(\mathcal X_{\mathcal W})
+
E_{\mathrm{IMU}}(\mathcal X_{\mathcal W}).
\end{aligned}
\label{eq:app_local_window}
\end{equation}
{Here, $\mathbf T_i^{\mathrm{LI}}=(\mathbf R_i,\mathbf p_i)$ maps the IMU frame to the LiDAR--inertial world frame; $\mathbf v_i$ is the world-frame velocity, and $\mathbf b_i^{g},\mathbf b_i^{a}$ are gyroscope and accelerometer biases. We transform scan points using these poses and calibrated LiDAR--IMU extrinsics, then group them into adaptive voxels. For each planar voxel $\ell\in\mathcal V$, let $\mathcal P_\ell$ contain its transformed scan points and fixed historical map points, with $n_\ell=|\mathcal P_\ell|$. Its mean, covariance, and contribution to the LiDAR objective are}
\begin{equation}
\begin{aligned}
\bar{\mathbf x}_\ell
&=
\frac{1}{n_\ell}
\sum_{\mathbf x\in\mathcal P_\ell}\mathbf x,\\
\mathbf C_\ell(\mathcal X_{\mathcal W})
&=
\frac{1}{n_\ell}
\sum_{\mathbf x\in\mathcal P_\ell}
(\mathbf x-\bar{\mathbf x}_\ell)
(\mathbf x-\bar{\mathbf x}_\ell)^\top,\\
E_{\mathrm{LiDAR}}(\mathcal X_{\mathcal W})
&=
\sum_{\ell\in\mathcal V}
\lambda_{\min}\!\left(
\mathbf C_\ell(\mathcal X_{\mathcal W})
\right).
\end{aligned}
\label{eq:app_lidar_plane_energy}
\end{equation}
{PCA defines a plane through $\bar{\mathbf x}_\ell$ whose normal is the eigenvector associated with $\lambda_{\min}$. This eigenvalue equals the mean squared point-to-plane distance. Minimizing it reduces point-cloud thickness and aligns scans to shared planes; the plane parameters are determined analytically from the current poses.}

{The IMU term constrains consecutive poses and velocities using preintegrated measurements, while accounting for bias changes:}
\begin{equation}
\begin{aligned}
\mathbf r^{I}_{i,i+1}
&=
\begin{bmatrix}
\operatorname{Log}\!\left(
\Delta\mathbf R_i^\top
\mathbf R_i^\top\mathbf R_{i+1}
\right)\\
\mathbf R_i^\top
\left(
\mathbf p_{i+1}-\mathbf p_i
-\mathbf v_i\Delta t_i
-\frac{1}{2}\mathbf g\Delta t_i^2
\right)-\Delta\mathbf p_i\\
\mathbf R_i^\top
\left(
\mathbf v_{i+1}-\mathbf v_i
-\mathbf g\Delta t_i
\right)-\Delta\mathbf v_i\\
\mathbf b_{i+1}^{g}-\mathbf b_i^{g}\\
\mathbf b_{i+1}^{a}-\mathbf b_i^{a}
\end{bmatrix},\\
E_{\mathrm{IMU}}(\mathcal X_{\mathcal W})
&=
\frac{1}{2}
\sum_{i=1}^{N-1}
\left\|
\mathbf r^{I}_{i,i+1}
\right\|_{\boldsymbol\Sigma_{i,i+1}^{-1}}^2.
\end{aligned}
\label{eq:app_imu_preintegration}
\end{equation}
{Here, $\Delta t_i=\tau_{i+1}-\tau_i$, and $\Delta\mathbf R_i$, $\Delta\mathbf p_i$, and $\Delta\mathbf v_i$ are bias-corrected preintegrated increments. The gravity vector $\mathbf g$ remains fixed after initialization. $\operatorname{Log}$ maps a rotation to its three-dimensional rotation vector, and $\boldsymbol\Sigma_{i,i+1}$ is the residual covariance, with $\|\mathbf r\|_{\boldsymbol\Sigma^{-1}}^2=\mathbf r^\top\boldsymbol\Sigma^{-1}\mathbf r$.}

\noindent\textbf{Metric position anchoring.}
{Once optimized LiDAR--inertial poses are finalized, we interpolate them at supported keyframe timestamps and convert them into camera-pose anchors:}
\begin{equation}
\mathbf T_j^{\mathrm M}
=
\mathbf T_{WW_{\mathrm L}}\,
\mathbf T^{\mathrm{LI},\star}(t_j)\,
\mathbf T_{BC}.
\label{eq:app_metric_anchor_conversion}
\end{equation}
{Here, $\mathbf T^{\mathrm{LI},\star}(t_j)$ is the interpolated IMU-to-world pose, $\mathbf T_{BC}$ maps camera coordinates to the IMU frame, and $\mathbf T_{WW_{\mathrm L}}$ rigidly aligns the LiDAR--inertial and visual world frames without scale fitting. Only keyframes bracketed by available finalized poses receive anchors. Their positions $\mathbf p_j^{\mathrm M}$ remain fixed during the subsequent visual position refinement, using the anchor prior defined in Section~\ref{sec:local_optimization}. Full pose anchors are retained for global correction and trajectory fusion; visual refinement does not modify the LiDAR--inertial window.}

\begin{figure}[!t]
    \centering
    \includegraphics[width=0.7\linewidth]{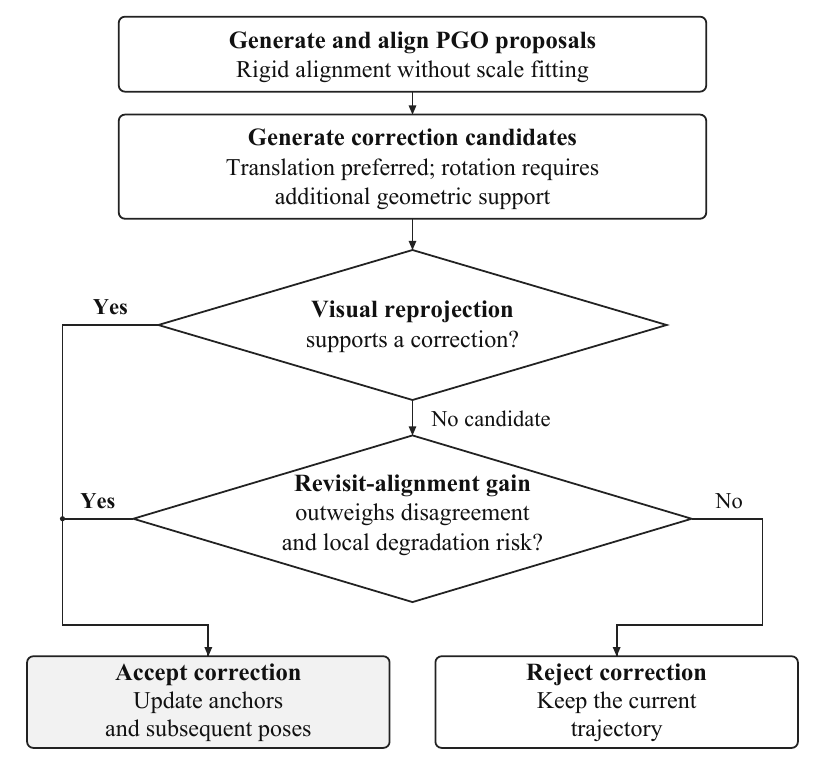}
    \caption{\textbf{Observation-supported loop correction.}
    Visual evidence is evaluated first, with geometric fallback
    when no visually supported candidate is selected.
    Rotation corrections require additional geometric support.}
    \label{fig:loop_correction_selection}
\end{figure}
\subsubsection{Observation-Supported Loop Correction}
\label{app:loop_correction}

{We obtain initial loop candidates and PGO proposals following~\citep{liu2026voxelslam}. Each proposal is rigidly aligned to the current metric trajectory without scale fitting and converted to the camera-pose convention. Given current anchors $\mathbf T_i^{\mathrm M}=(\mathbf R_i^{\mathrm M},\mathbf p_i^{\mathrm M})$ and aligned proposals $\widetilde{\mathbf T}_i^{\mathrm M}$, we independently control translation and rotation corrections:}
\begin{equation}
\begin{aligned}
\mathbf p_i^{\mathrm M,\alpha}
&=
\mathbf p_i^{\mathrm M}
+
\alpha\left(
\widetilde{\mathbf p}_i^{\mathrm M}
-
\mathbf p_i^{\mathrm M}
\right),\\
\mathbf R_i^{\mathrm M,\beta}
&=
\mathbf R_i^{\mathrm M}
\operatorname{Exp}\!\left(
\beta\operatorname{Log}\!\left(
(\mathbf R_i^{\mathrm M})^\top
\widetilde{\mathbf R}_i^{\mathrm M}
\right)
\right).
\end{aligned}
\label{eq:app_loop_correction_gains}
\end{equation}
{The gains $\alpha,\beta$ are drawn from a predefined discrete set $\mathcal G\subset[0,1]$: zero preserves the current estimate, while one applies the full proposed correction.}

{Figure~\ref{fig:loop_correction_selection} summarizes candidate selection using available observations. Supported visual reprojection evidence is checked first. If no candidate is selected, geometric gains must outweigh disagreement across observation subsets and local degradation risk, with visual corroboration where available. Both branches favor translation-only corrections; rotation requires additional geometric benefit relative to its risk. Accepted corrections update the supported anchors and propagate to subsequent metric poses; rejection preserves the current accepted trajectory.}

\subsubsection{Metric-Anchored Visual Motion Fusion}
\label{app:trajectory_fusion}

{We apply Equation~\ref{eq:global_visual_motion_fusion} between consecutive accepted metric anchors with visual support, using the following interpolation and update rules:}

{\textbf{(1) Interpolation.}
Metric anchors and visual endpoint poses are interpolated at the same sensor timestamps using linear position interpolation and shortest-arc quaternion SLERP. An interval is fused only after both anchor estimates become available.}

{\textbf{(2) Anchor updates.}
Accepted loop corrections trigger recomputation of the affected fused poses using the updated anchors and the stored visual trajectory. The visual translation residual remains in world coordinates, while the visual rotation residual is composed with the interpolated metric orientation.}

{\textbf{(3) Boundary handling.}
Before the right anchor becomes available, poses follow available LiDAR--IMU odometry relative to the latest accepted anchor and are revised upon finalization. A camera-only tail at sequence end follows visual relative motion from the last supported fused pose. Startup frames without metric support retain their initial pose estimates.}

\subsubsection{Optional Offline Hierarchical Pose Refinement}
\label{app:offline_hba}

{After streaming and loop correction, we initialize hierarchical LiDAR bundle adjustment~\citep{liu2023hba} from the accepted camera trajectory, converted to scan-time IMU poses using calibrated extrinsics and coordinate transforms. Hierarchical optimization refines LiDAR alignment within and across submaps, and the resulting pose increments are transferred back to camera timestamps. }

\subsection{Geometry-Aware Gaussian Reconstruction Details}
\label{app:gaussian_reconstruction}

\subsubsection{Streaming Gaussian Reconstruction}
\label{app:streaming_gaussian}

\noindent\textbf{Representation and initialization.}
{Following~\citep{jiang2026planing}, each triangle carries learnable vertices, opacity, and appearance features, and supports differentiable depth and normal rendering. Associated Gaussians model appearance, with their centers parameterized by the triangle barycenter plus learnable 3D offsets. For each mapping keyframe, positive differences between observed and rendered Laplacian-of-Gaussian responses identify unresolved image detail, while depth-adaptive spatial filtering suppresses redundant primitives. LiDAR-derived depth and normals initialize triangle positions and orientations. Associated Gaussians start at the triangle barycenter, with appearance initialized from RGB observations.}

\noindent\textbf{Rendering losses.}
{With camera poses fixed during each mapping update, we jointly optimize geometry and appearance using}
\begin{equation}
\begin{aligned}
\mathcal L_{\mathrm{stream}}
&=
\mathcal L_{\mathrm{photo}}
+
\lambda_d
\left\|
\widehat D^{\mathrm T}-D
\right\|_1
+
\lambda_n
\left\|
\widehat{\mathbf N}^{\mathrm T}-\mathbf N
\right\|_1
\\
&\quad+
\lambda_{\mathrm{inv}}
\left\|
(\widehat D^{\mathrm G})^{-1}-D^{-1}
\right\|_1
+
\lambda_o\mathcal L_{\mathrm{opacity}},
\\
\mathcal L_{\mathrm{photo}}
&=
(1-\lambda_s)
\left\|
\widehat{\mathbf I}^{\mathrm G}-\mathbf I
\right\|_1
+
\lambda_s
\left[
1-\operatorname{SSIM}
(\widehat{\mathbf I}^{\mathrm G},\mathbf I)
\right].
\end{aligned}
\label{eq:app_streaming_loss}
\end{equation}
{Here, $\mathbf I$ is the observed RGB image, and $D,\mathbf N$ are LiDAR-derived depth and normal targets. Superscripts $\mathrm T$ and $\mathrm G$ denote triangle and Gaussian renderings, respectively. The norms denote mean absolute errors over valid pixels, and $\mathcal L_{\mathrm{opacity}}$ regularizes triangle opacity following~\citep{jiang2026planing}. Depth and normal losses constrain triangle geometry, inverse-depth supervision constrains Gaussian placement, and photometric supervision refines appearance while also updating geometry through the triangle--Gaussian association. The $\lambda$ coefficients balance these terms.}

\noindent\textbf{Pose updates.}
{When a source-keyframe pose changes, we apply its rigid correction to the associated triangle vertices and Gaussian centers, rotating Gaussian offsets and orientations accordingly. This keeps the map consistent with the corrected camera trajectory.}

\noindent\textbf{CPU--GPU memory management.}
{To reduce GPU memory usage as the scene grows, we offload inactive primitives and their optimizer states to CPU memory and reload them when their source keyframes become relevant. Triangle--Gaussian associations and optimizer states are preserved across transfers, allowing optimization to resume without reinitialization. Pose corrections update both CPU- and GPU-resident primitives to maintain consistency throughout the map.}

\subsubsection{Geometry-Aware Gaussian Refinement}
\label{app:gaussian_refinement}

\noindent\textbf{Diagnostic evidence.}
{We refine the exported Gaussians directly, keeping camera poses and exposure transforms fixed. Motion-compensated LiDAR projections provide depth and local normals for diagnosing regions with high RGB rendering error. For a Gaussian $g$, we evaluate}
\begin{equation}
e_f(\mathbf u)
=
\frac{1}{3}
\left\|
\widehat{\mathbf I}_f^{\mathrm G}(\mathbf u)
-
\mathbf I_f(\mathbf u)
\right\|_1,
\qquad
E_g
=
\frac{1}{|\mathcal O_g|}
\sum_{f\in\mathcal O_g}
e_f(\mathbf u_{fg}),
\label{eq:app_repair_image_error}
\end{equation}
{where $\mathbf u_{fg}$ is the projected Gaussian center and $\mathcal O_g$ contains eligible training views with nearby LiDAR support, sufficient rendered opacity, and no foreground occlusion. Target positions and normals must agree across views. An observation is geometrically consistent when both its axial-depth error and point-to-plane distance are below the depth-dependent tolerance $\tau(z)$. It is inconsistent when its depth error lies between $2\tau(z)$ and an upper bound $d_{\max}$ and its RGB error exceeds $\tau_r$. Let $\eta_g^{+}$ and $\eta_g^{-}$ denote the corresponding fractions of eligible views.}

\noindent\textbf{Repair operations.}
{We apply three repair operations based on rendering error, LiDAR depth agreement, and Gaussian coverage.}

\noindent\textbf{(1) Split oversized, geometrically consistent Gaussians.}
{We select Gaussians with $E_g>\tau_s$,
$\eta_g^{+}>\eta_{\min}$, projected radius above $r_{\min}$,
and largest scale $s_g>s_{\min}$. The corresponding principal
axis $\mathbf a_g$ must be approximately tangential to the
aggregated LiDAR normal $\mathbf n_g$, satisfying
$|\mathbf a_g^\top\mathbf n_g|<\tau_{\mathrm{tan}}$.
Each selected parent is replaced by two children:}
\begin{equation}
\mathbf c_{g,\pm}
=
\mathbf c_g
\pm\frac{s_g}{\sqrt{2}}\mathbf a_g,
\qquad
s_{g,\pm}^{(a)}
=
\frac{s_g}{\sqrt{2}},
\qquad
\omega_{g,\pm}
=
\frac{\omega_g}{\sqrt{2}},
\label{eq:app_gaussian_split}
\end{equation}
{where $s_{g,\pm}^{(a)}$ denotes the child scale along
$\mathbf a_g$ and $\omega_g$ is the parent opacity.
All other attributes are inherited.}

\noindent\textbf{(2) Replace Gaussians with persistent depth disagreement.}
{We select Gaussians with $E_g>\tau_r$ and
$\eta_g^{-}>\eta_{\min}$. Each center is projected onto
LiDAR-supported tangent planes, and the projected positions
are aggregated into a target $\mathbf q_g$. We reinitialize
the Gaussian at $\mathbf q_g$, aligning its thin axis with
the aggregated normal and initializing compact tangential
scales and appearance from exposure-corrected training colors.}

\noindent\textbf{(3) Insert Gaussians in underrepresented regions.}
{For a pixel with valid LiDAR depth $z$ and rendering error
$e_f(\mathbf u)>\tau_a$, insertion requires insufficient
coverage and visibility, respectively:}
\begin{equation}
\begin{aligned}
&\widehat A_f(\mathbf u)<a_{\mathrm{cov}}
\quad\text{or}\quad
\widehat D_f^{\mathrm G}(\mathbf u)>z+2\tau(z),\\
&\widehat A_f(\mathbf u)<a_{\mathrm{empty}}
\quad\text{or}\quad
\widehat D_f^{\mathrm G}(\mathbf u)\geq z-\tau(z).
\end{aligned}
\label{eq:app_gaussian_insertion}
\end{equation}
{Both conditions must hold. Here, $\widehat A_f$ is
accumulated rendered opacity; the second condition prevents
insertion behind an already modeled foreground surface.
Candidates must exhibit consistent normals and colors
across multiple views. We retain the highest-error observation
per qualifying spatial bin and reject candidates near
existing Gaussian centers. LiDAR depth initializes position,
local normals determine orientation, and training RGB
observations initialize appearance.}

\noindent\textbf{Refinement objective.}
{After repair, we optimize Gaussian centers, scales, rotations,
opacities, and spherical-harmonic coefficients using}
\begin{equation}
\begin{aligned}
\mathcal L_{\mathrm{refine}}
&=
\mathcal L_{\mathrm{photo}}
+
\lambda_e\mathcal L_{\mathrm{edge}}
+
\lambda_p\mathcal L_{\mathrm{plane}},\\
\mathcal L_{\mathrm{plane}}
&=
\frac{1}{|\mathcal A|}
\sum_{g\in\mathcal A}
\operatorname{SmoothL1}\!\left(
\frac{
\mathbf n_g^\top(\mathbf c_g-\mathbf q_g)
}{
\sigma_p
}
\right).
\end{aligned}
\label{eq:app_gaussian_refinement_loss}
\end{equation}
{The photometric term follows
Equation~\ref{eq:app_streaming_loss}.
$\mathcal L_{\mathrm{edge}}$ compares signed horizontal and
vertical Sobel derivatives of image luminance at full and
half resolution, using a Charbonnier penalty and detached
residual-based weights.
The set $\mathcal A$ includes newly created or reinitialized
Gaussians with valid LiDAR targets, including supported
split children, replacements, and insertions.
Each target $(\mathbf q_g,\mathbf n_g)$ is held fixed during
optimization, and $\sigma_p$ sets the normal-displacement scale.
The plane term is zero when $\mathcal A$ is empty.
It constrains motion along the LiDAR normal while allowing
tangential adjustment.}

\noindent\textbf{Repair schedule.}
{Refinement runs for 15,000 steps.
All three repair operations are applied every 1,000 steps
from step 2,000 through 8,000.
Unchanged primitives retain their Adam states, while
split children, replacements, and insertions receive
reset optimizer moments.}

\begin{figure}[!t]
    \centering
    \includegraphics[width=\linewidth]{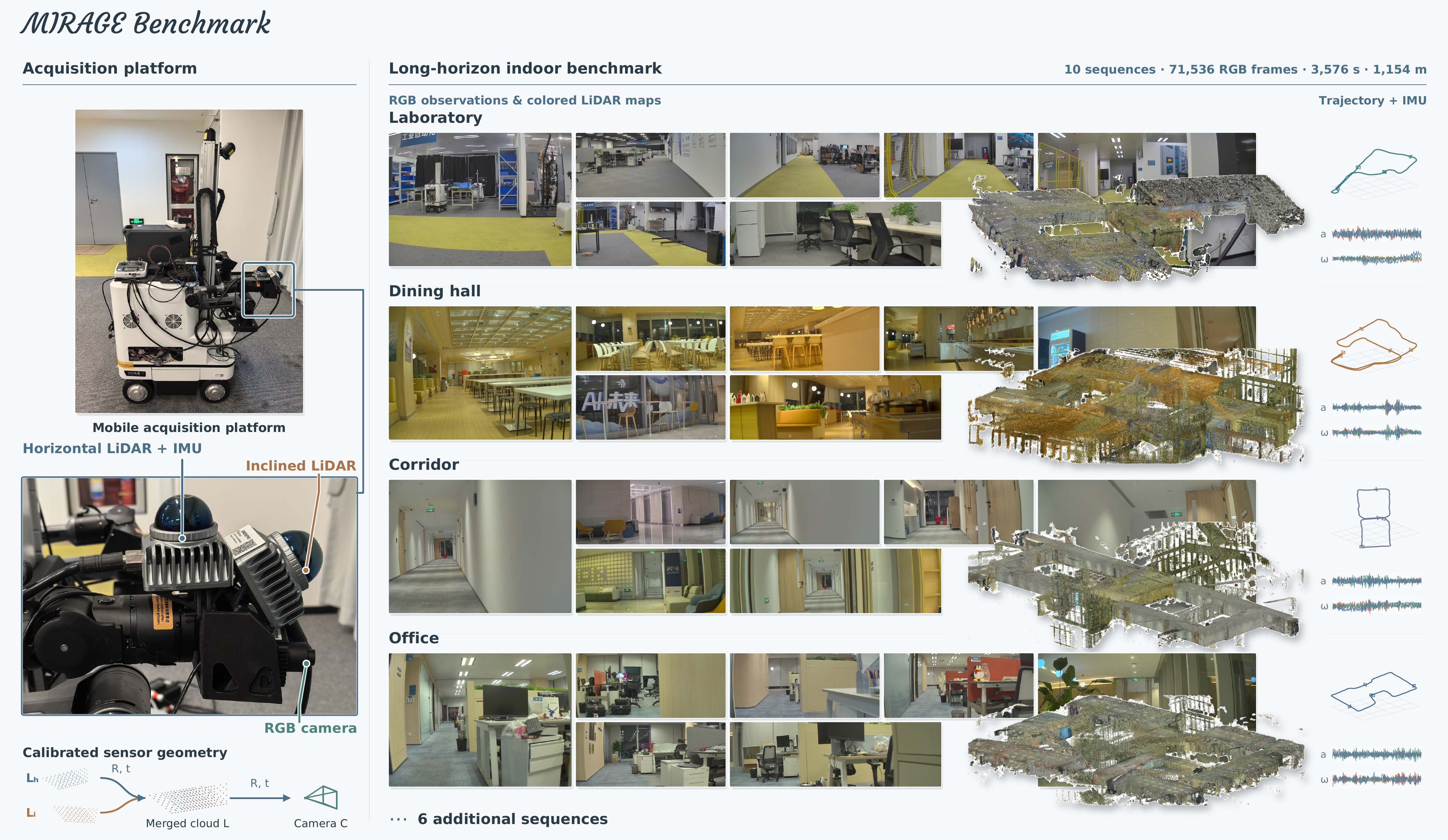}
    \caption{\textbf{Acquisition platform and dataset overview.}
    \textbf{Left:} the synchronized RGB--LiDAR--IMU platform, sensor layout, and calibrated sensor transforms.
    \textbf{Right:} RGB images, colored LiDAR maps, MIRAGE reference trajectories, and IMU signals from four representative sequences spanning laboratories, dining areas, corridors, and offices.}
    \label{fig:mirage_benchmark_supp}
\end{figure}

\begin{figure}[!t]
    \centering
    \includegraphics[width=\linewidth]{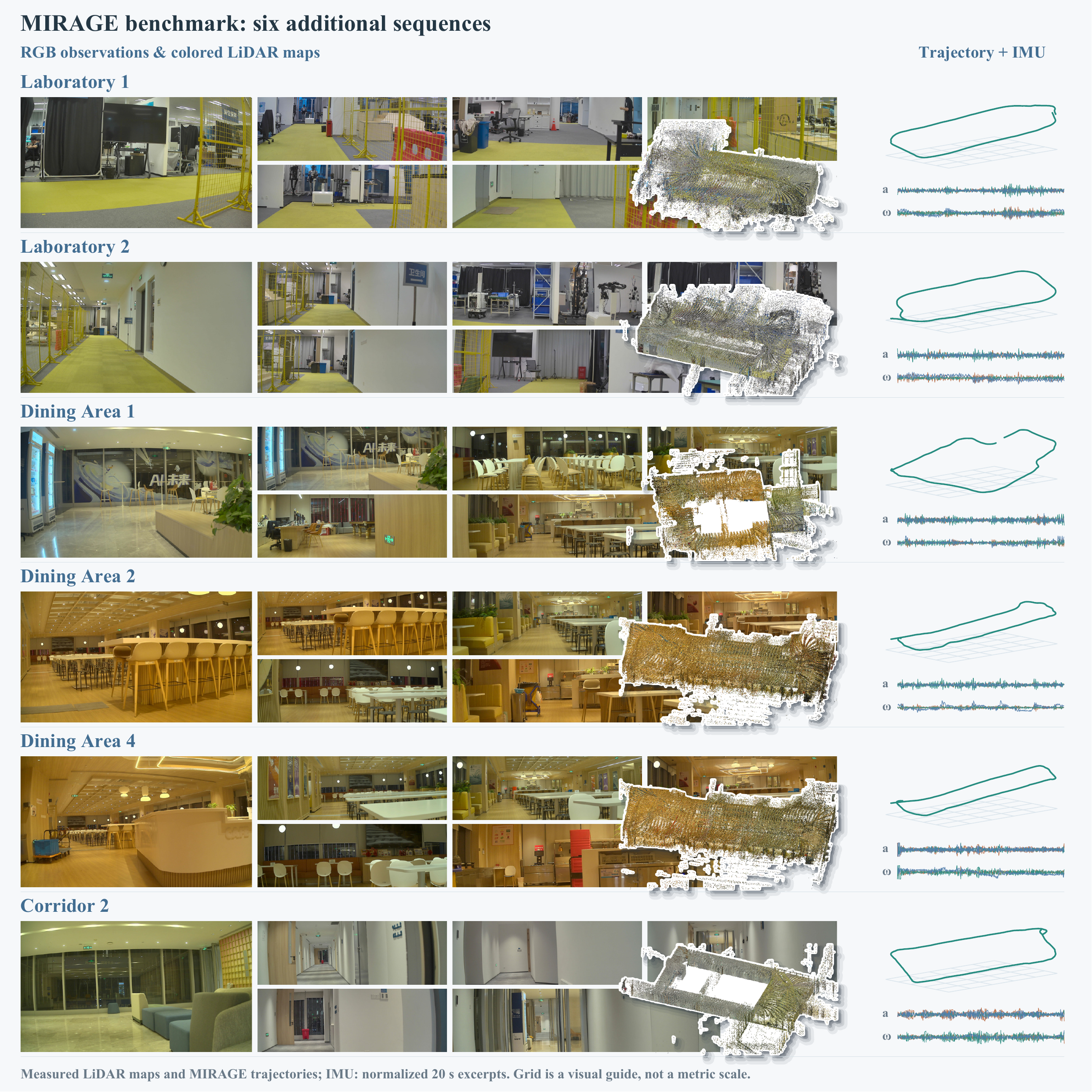}
    \caption{\textbf{Six additional MIRAGE benchmark sequences.}
    Complementing Figure~\ref{fig:mirage_benchmark_supp}, each row shows RGB images and colored LiDAR maps (left), alongside MIRAGE reference trajectories and IMU signals (right).
    IMU plots show normalized 20-second excerpts; trajectory grids are illustrative and do not indicate metric scale.}
    \label{fig:mirage_benchmark_supp2}
\end{figure}

\subsection{Data Acquisition and Benchmark}
\label{app:data_acquisition}

\subsubsection{Sensor Suite and Robot Platform}
\label{app:sensor_platform}
{Our data collection platform (Figure~\ref{fig:mirage_benchmark_supp}) integrates an industrial RGB camera (\texttt{SENSING TRD1-G2A GMSL2}), two Livox MID-360 LiDARs with built-in IMUs, and an onboard computer.
Mounted on an AgileX mobile robot, the platform supports teleoperated data collection and online scene reconstruction. Each LiDAR has a horizontal field of view (FoV) of $360^\circ$ and a vertical FoV of $-7^\circ$ to $52^\circ$. 
Following RISED~\citep{jiang2025rised}, we mount one LiDAR horizontally and the other at an incline to broaden spatial coverage while avoiding mutual occlusion.
The horizontal LiDAR captures distant structures to constrain pose estimation, while the inclined LiDAR densely samples nearby surfaces for detailed mapping. Inertial measurements are provided by the horizontal LiDAR's built-in IMU.}

{The camera is connected to an NVIDIA Thor device. It captures images at a resolution of $2880 \times 1860$ pixels, with a horizontal FoV of $64^\circ$, a vertical FoV of $40^\circ$, and an
acquisition rate of $20\,\mathrm{Hz}$.
Images and their associated timestamps are streamed from Thor to an
onboard industrial computer through ZeroMQ. The industrial computer
is equipped with an NVIDIA RTX 4090 GPU for SLAM and scene reconstruction.}

\subsubsection{Calibration and Time Synchronization}
\label{app:calibration_sync}

{We estimate the inter-LiDAR extrinsics using GICP~\citep{chen2022direct}. Initialized from the assembly drawings, a shared rigid transform is jointly refined using point clouds collected in five distinct scenes with the platform stationary.}
{LiDAR--camera extrinsics are estimated using FAST-Calib~\citep{zheng2026fast}. We acquire paired images and LiDAR recordings of a calibration target at multiple positions, orientations, and distances to provide complementary geometric constraints. We qualitatively validate the estimated extrinsics by projecting RGB colors onto the LiDAR point cloud and visually inspecting the colorization for misalignment artifacts.}

{Time synchronization combines Precision Time Protocol (PTP) clock
synchronization with external camera triggering. A Xingqi XQ500
synchronization unit provides the common time reference and camera
trigger signals. The LiDAR clocks synchronize to the PTP reference
over Ethernet. Each image is timestamped with its corresponding camera trigger time.     We verify image timestamps by imaging a reference clock display
and comparing the displayed times with the recorded timestamps,
with observed discrepancies below $10\,\mathrm{ms}$.}

\subsubsection{Collection Protocol and Dataset Statistics}
\label{app:dataset_collection}
{Our dataset comprises 10 sequences across four indoor scene categories: laboratories, dining areas, an office, and long corridors. Laboratories and dining areas contain numerous objects and reflective surfaces, challenging detailed geometry recovery and appearance modeling. Long, low-texture corridors challenge visual tracking and long-term trajectory consistency. We collected all sequences through low-speed teleoperation to limit motion blur and manually inspected the RGB images to remove blurred frames.}

{The dataset contains 71,536 RGB images at $2880\times1860$ resolution and 35,770 LiDAR frames, together with synchronized IMU measurements, spanning 59.6 minutes and approximately 1.15~km. Individual sequences range from 1.77 to 11.11 minutes, with trajectories approaching 200~m. Table~\ref{tab:dataset_statistics} reports per-sequence statistics, with sequence names matching the benchmark evaluation. Figure~\ref{fig:mirage_benchmark_supp} presents the acquisition platform and four representative sequences spanning all scene categories, while Figure~\ref{fig:mirage_benchmark_supp2} shows the remaining six sequences. Together, they illustrate scene appearance, LiDAR maps, reference trajectories, and inertial measurements across the complete dataset.}

\begin{table}[!t]
    \caption{Per-sequence acquisition statistics of the MIRAGE benchmark.}
    \label{tab:dataset_statistics}
    \centering
    \resizebox{0.65\linewidth}{!}{%
        \begin{tabular}{lrrrr}
            \toprule
            Sequence
            & \makecell{\# RGB\\Images}
            & \makecell{\# LiDAR\\Frames}
            & \makecell{Duration\\(min)}
            & \makecell{Trajectory\\Length (m)} \\
            \midrule
            Laboratory 1  & 4,112  & 2,056 & 3.42  & 54.57  \\
            Laboratory 2  & 4,470  & 2,235 & 3.72  & 61.02  \\
            Laboratory 3  & 13,336 & 6,668 & 11.11 & 168.46 \\
            \midrule
            Dining Area 1 & 4,430  & 2,215 & 3.69  & 112.69 \\
            Dining Area 2 & 2,126  & 1,063 & 1.77  & 64.65  \\
            Dining Area 3 & 11,667 & 5,834 & 9.72  & 181.52 \\
            Dining Area 4 & 6,397  & 3,199 & 5.33  & 65.37  \\
            \midrule
            Corridor 1    & 11,729 & 5,865 & 9.77  & 165.39 \\
            Corridor 2    & 8,070  & 4,035 & 6.72  & 85.57  \\
            \midrule
            Office 1      & 5,199  & 2,600 & 4.33  & 195.15 \\
            \midrule
            \textbf{Total}
            & \textbf{71,536}
            & \textbf{35,770}
            & \textbf{59.60}
            & \textbf{1,154.40} \\
            \bottomrule
        \end{tabular}%
    }
\end{table}
 
\begin{figure}[!t]
    \centering
    \includegraphics[width=\linewidth]{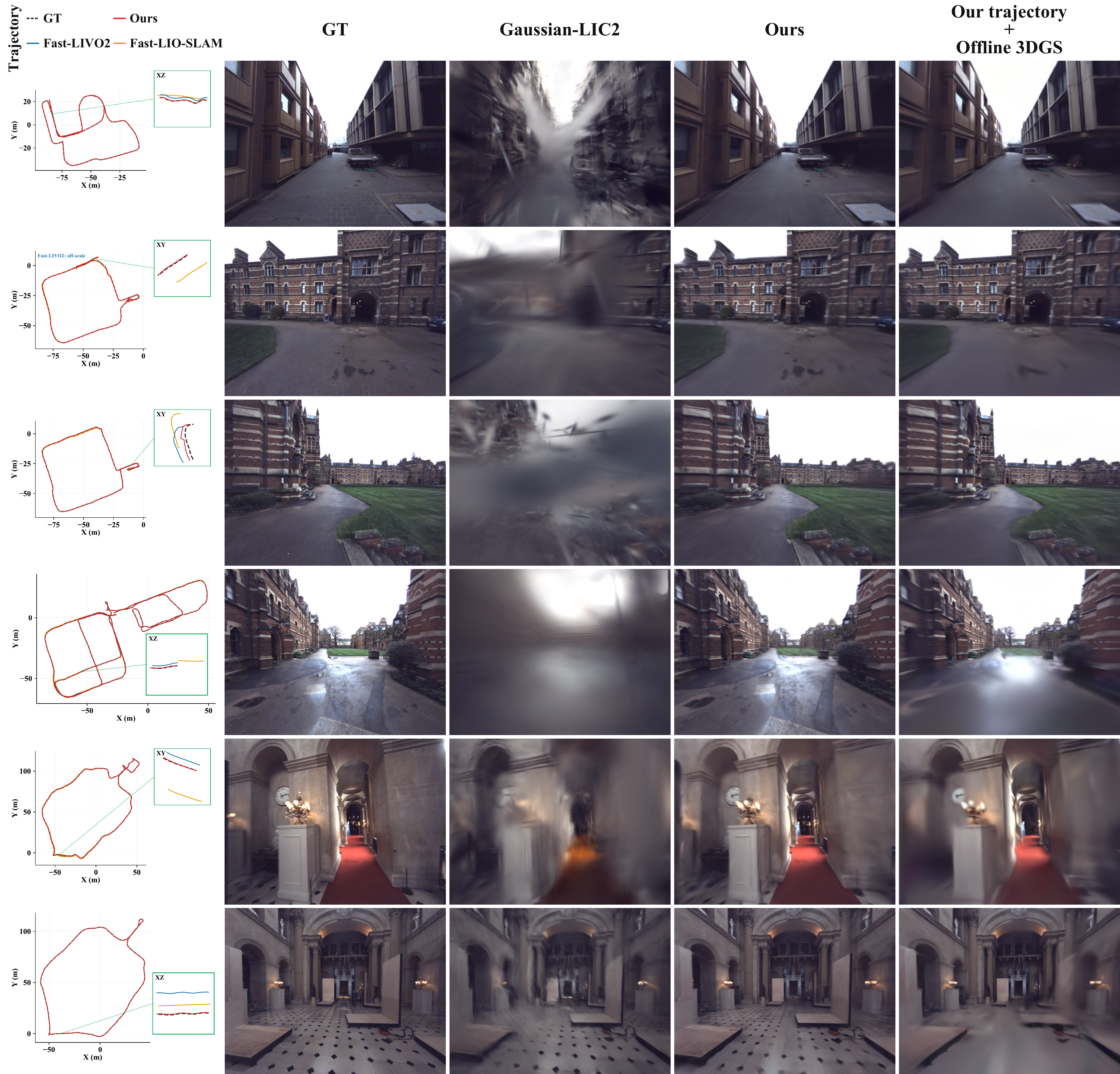}
    \caption{\textbf{Additional qualitative comparisons on Oxford Spires.} Rows, from top to bottom: Observatory~01, Keble~02, Keble~03, Keble~04, Blenheim~01, and Blenheim~02. Left: SE(3)-aligned XY trajectories, with green boxes and connectors indicating local views in the labeled planes at equal axis scales. Fast-LIVO2 is off-scale on Keble~02. Right: held-out ground-truth images, Gaussian-LIC2, MIRAGE, and offline 3DGS using MIRAGE trajectories.}
    \label{fig:supp_qualitative_oxford}
\end{figure}

\begin{figure}[!t]
    \centering
    \includegraphics[width=\linewidth]{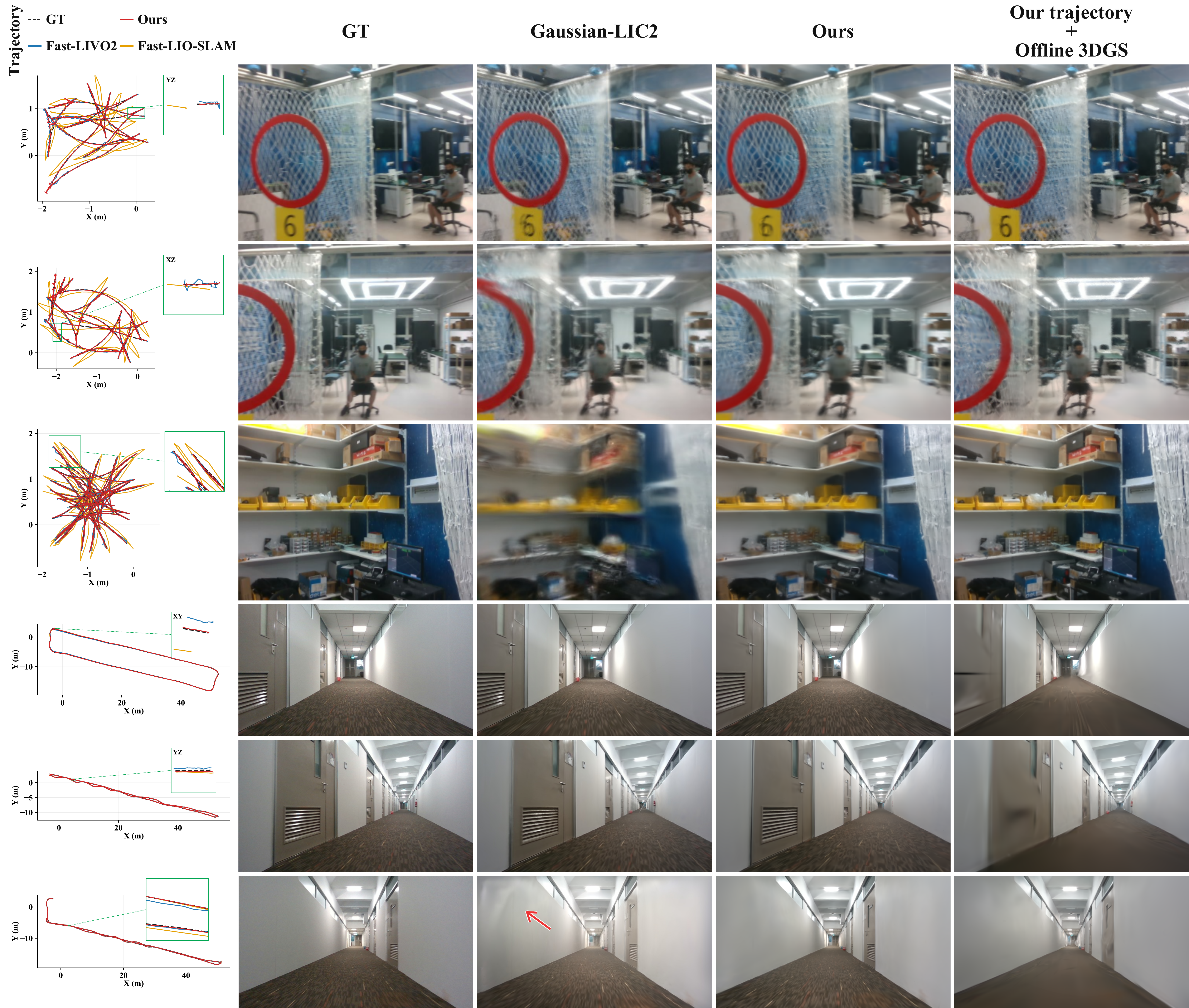}
    \caption{\textbf{Additional qualitative comparisons on M2DGR and INS.} Rows, from top to bottom: room\_01, room\_02, room\_03, indoor\_sq1, indoor\_sq3, and indoor\_sq4. Columns and green trajectory insets follow Figure~\ref{fig:supp_qualitative_oxford}. The arrow highlights a wall artifact in the Gaussian-LIC2 reconstruction of indoor\_sq4.}
    \label{fig:supp_qualitative_m2dgr_ins}
\end{figure}

\subsection{Implementation and Evaluation Protocols}
\label{app:implementation_evaluation}

\noindent\textbf{Implementation.}
{All experiments use NVIDIA RTX 4090 GPUs. LiDAR--inertial bundle adjustment uses a ten-scan sliding window. Gaussian refinement runs for 15,000 steps with camera poses and exposure transforms fixed, following the repair schedule in Appendix~\ref{app:gaussian_refinement}. Offline 3DGS baselines are trained for 30,000 steps.}

\noindent\textbf{Evaluation protocols.}
{Trajectory evaluation reports translational ATE RMSE after SE(3) alignment for metric methods and Sim(3) alignment for vision-only methods. Each trajectory is evaluated at timestamps with valid reference associations. Reported means weight sequences equally. Pose-stage comparisons use matched timestamps within each run; offline HBA results are reported separately rather than treated as a paired stage ablation.}

{Rendering evaluation holds out every eighth image from Gaussian optimization. On Oxford Spires, M2DGR, and INS, offline 3DGS uses fixed MIRAGE online poses. MIRAGE rendering results use the Gaussian maps after refinement, without HBA pose updates; Gaussian-LIC2 results use its online maps. Comparisons exclude Observatory~02, Church~03, and indoor\_sq2 because Gaussian-LIC2 did not complete reconstruction.}

{On the MIRAGE benchmark, visual baselines use one-in-twelve sampling of the same preprocessed RGB sequences, without LiDAR, IMU, or reference-pose inputs. Each method supplies only its predicted poses to downstream 3DGS. These poses remain fixed during training, and Gaussians are initialized from training-view correspondences using a common triangulation procedure. The common-frame evaluation uses shared training and test images. MASt3R-SLAM is evaluated on registered keyframes, and its rendering results are excluded from common-frame rankings. Ours (reference) uses the same 3DGS procedure with MIRAGE reference poses. Stream PSNR reports the complete MIRAGE pipeline after refinement on its own held-out images, rather than the shared stride-12 test set.}

{For the visual ablation, camera poses are obtained by interpolating LiDAR--inertial poses to image timestamps and applying calibrated extrinsics; RGB supervision for Gaussian reconstruction is retained. Photometric and geometry-aware refinement start from the same streaming map and share the optimization budget, fixed poses and exposure, renderer, and test views. Photometric refinement uses only $\mathcal L_1$ and SSIM losses with a fixed primitive count.}

\subsection{Additional Experimental Results}
\label{app:additional_results}

\paragraph{Per-Sequence Pose Stages.}
Table~\ref{tab:pose_stages_detailed} gives the per-sequence values underlying the online averages in Table~\ref{tab:ablation_summary}(a). Offline (+HBA) averages use the same scene set and the unrounded source results for Table~\ref{tab:main_pose_results}.

\begin{table}[t]
\centering
\caption{Stage-wise translational ATE RMSE (cm, $\downarrow$), with independent fixed-scale SE(3) alignment on common within-run timestamps. Tracking denotes saved visual inputs to metric refinement; Local and Global denote the no-loop and corrected camera trajectories.}
\label{tab:pose_stages_detailed}
\begingroup
\small
\setlength{\tabcolsep}{3pt}
\resizebox{\linewidth}{!}{%
\begin{tabular}{l*{15}{c}}
\toprule
& \multicolumn{8}{c}{\textit{Oxford Spires}} & \multicolumn{3}{c}{\textit{M2DGR}} & \multicolumn{4}{c}{\textit{INS}} \\
\cmidrule(lr){2-9}\cmidrule(lr){10-12}\cmidrule(lr){13-16}
Stage & \makecell[c]{Observatory\\01} & \makecell[c]{Observatory\\02} & \makecell[c]{Keble\\02} & \makecell[c]{Keble\\03} & \makecell[c]{Keble\\04} & \makecell[c]{Blenheim\\01} & \makecell[c]{Blenheim\\02} & \makecell[c]{Church\\03} & room\_01 & room\_02 & room\_03 & \makecell[c]{indoor\\sq1} & \makecell[c]{indoor\\sq2} & \makecell[c]{indoor\\sq3} & \makecell[c]{indoor\\sq4} \\
\midrule
Tracking & 69.739 & 97.263 & 121.833 & 77.076 & 57.736 & 70.032 & 149.269 & 15.960 & 31.694 & 20.781 & 13.128 & 48.784 & 63.709 & 30.351 & 32.431 \\
Local & 4.242 & 4.695 & 5.083 & 7.162 & 10.725 & 14.654 & 14.473 & 1.239 & 1.828 & 1.926 & 1.971 & 13.148 & 5.468 & 5.172 & 4.433 \\
Global & 3.429 & 3.559 & 3.595 & 7.125 & 9.611 & 4.495 & 3.265 & 1.154 & 1.873 & 2.179 & 2.148 & 5.666 & 5.477 & 5.198 & 4.441 \\
\bottomrule
\end{tabular}%
}
\endgroup
\end{table}

Figures~\ref{fig:supp_qualitative_oxford} and~\ref{fig:supp_qualitative_m2dgr_ins} extend the trajectory and held-out rendering comparisons to all 12 sequences in Table~\ref{tab:main_rendering_results}. Observatory~02, Church~03, and indoor\_sq2 are excluded because Gaussian-LIC2 failed on these sequences.

\subsection{Limitations}
\label{app:limitations}
{While MIRAGE achieves accurate metric pose estimation and high-fidelity reconstruction, several limitations remain. First, our benchmark focuses on static indoor scenes and does not assess the impact of dynamic objects on localization and reconstruction. Second, our data acquisition and reconstruction pipelines lack dedicated treatment of transparent and semi-transparent objects, which can compromise geometric accuracy and rendering quality. These limitations motivate future work on extending the benchmark to dynamic scenes and improving reconstruction of challenging materials.}

\end{document}